\documentclass[journal,11pt,twocolumn]{IEEEtran} 

\ifCLASSOPTIONcompsoc
\usepackage[normalsize,sf,SF]{subfigure}
\else
\usepackage{subfigure}
\fi
\usepackage{cite}
\usepackage{amsmath,amssymb,bbm}
\usepackage{pifont}
\usepackage{url,booktabs}
\usepackage{graphicx}
\graphicspath{{../}{../LaCCAN/Abraao/ImgFig/}}
\usepackage[ugly]{units}
\usepackage{multirow}
\usepackage{algorithm,algpseudocode}

\newtheorem{LEO}{LEO}

\newtheorem{PEO}{PEO}

\newtheorem{Lemma}[LEO]{Lemma}
\newtheorem{Proposition}[PEO]{Proposition}

\begin{document}

\title{Analytic Expressions for Stochastic Distances Between Relaxed Complex Wishart Distributions}

\author{Alejandro C.\ Frery,~\IEEEmembership{Member},
Abra\~ao D.\ C.\ Nascimento, and
Renato J.\  Cintra,~\IEEEmembership{Senior Member}
\thanks{This work was supported by CNPq, Fapeal and FACEPE, Brazil.}
\thanks{
A.\ C.\ Frery is with the Instituto de Computa\c c\~ao, Universidade Federal de Alagoas, BR 104 Norte km 97, 57072-970, Macei\'o, AL, Brazil, email: acfrery@gmail.com}
\thanks{A.\ D.\ C.\ Nascimento and R.\ J.\ Cintra are with the Departamento de Estat\'istica, Universidade Federal de Pernambuco, Cidade Universit\'aria, 50740-540, Recife, PE, Brazil, e-mails: abraao.susej@gmail.com and rjdsc@de.ufpe.br}}

\markboth{IEEE Transactions on Geoscience and Remote Sensing}%
{Frery \MakeLowercase{\textit{et~al.}}: Analytic Wishart Distances}

\maketitle

\begin{abstract}
The scaled complex Wishart distribution is a widely used model for multilook full polarimetric SAR data whose adequacy has been attested in the literature.
Classification, segmentation, and image analysis techniques which depend on this model have been devised, and many of them employ some type of dissimilarity measure.
In this paper we derive analytic expressions for four stochastic distances between relaxed scaled complex Wishart distributions in their most general form and in important particular cases.
Using these distances, inequalities are obtained which lead to new ways of deriving the Bartlett and revised Wishart distances.
The expressiveness of the four analytic distances is assessed with respect to the variation of parameters. 
Such distances are then used for deriving new tests statistics, which are proved to have asymptotic chi-square distribution.
Adopting the test size as a comparison criterion, a sensitivity study is performed by means of Monte Carlo experiments suggesting that the Bhattacharyya statistic outperforms all the others.
The power of the tests is also assessed.
Applications to actual data illustrate the discrimination and homogeneity identification capabilities of these distances.
\end{abstract}
\begin{IEEEkeywords}
statistics, image analysis, information theory, polarimetric radar, contrast measures.
\end{IEEEkeywords}

\section{Introduction}

\IEEEPARstart{P}{olarimetric} Synthetic Aperture Radar~(PolSAR) devices transmit orthogonally polarized pulses towards a target, and the returned echo is recorded with respect to each polarization.
Such remote sensing apparatus provides the means for a better capture of scene information when compared to its univariate counterpart, namely the conventional SAR technology, and complementary information with respect to other remote sensing modalities~\cite{LeePottier2009PolarimetricRadarImaging,Lopez-MartinezFabregasPottier2005}.

PolSAR can achieve high spatial resolution due to its coherent processing of the returned echoes~\cite{OliverandQuegan1998}.
Being multichanneled by design, PolSAR also allows individual characterization of the targets in various channels. 
Moreover, it enables the identification of covariance structures among channels.

Resulting images from coherent systems are prone to a particular interference pattern called speckle~\cite{OliverandQuegan1998}.
This phenomenon can seriously affect the interpretation of PolSAR imagery~\cite{Lopez-MartinezFabregasPottier2005}.
Thus, specialized signal analysis techniques are usually required. 

Segmentation~\cite{BeaulieuTouzi2004}, classification~\cite{KerstenandLeeandAinsworth2005}, boundary detection~\cite{PolarimetricSegmentationBSplinesMSSP,GambiniandMejailandJacobo-BerllesandFrery}, and change detection~\cite{IngladaMercier2007} techniques often employ dissimilarity measures for data discrimination. 
Such measures have been used to quantify the difference between image regions, and are often called `contrast measures'.
The analytical derivation of contrast measures and their properties is an important venue for image understanding.
Methods based on numerical integration have several disadvantages with respect to closed formulas, such as lack of convergence of the iterative procedures, and high computational cost.
Stochastic distances between models for PolSAR data often require dealing with integrals whose domain is the set of all positive definite Hermitian matrices. 

Goudail and R\'efr\'egier~\cite{Goudail2004} applied stochastic measures to characterize the performance of target detection and segmentation algorithms in PolSAR image processing.
In that study, both Kullback-Leibler and Bhattacharyya distances were considered as tools for quantifying the dissimilarity between circular complex Gaussian distributions. 
The Bhattacharyya measure was reported to possess better contrast capabilities than the Kullback-Leibler measure.
However, the statistical properties of the measures were not explicitly considered in that work.

Erten \textit{et al}.~\cite{Ertenetal2012} derived a ``coherent similarity'' between PolSAR images based on the mutual information.
Morio \textit{et al}.~\cite{MorioRefregierGoudailFernandezDupuis2009} applied the Shannon entropy and Bhattacharyya distance for the characterization of polarimetric interferometric SAR images. 
They decomposed the Shannon entropy into the sum of three terms with physical meaning.

PolSAR theory prescribes that the returned (backscattered) signal of distributed targets is adequately represented by its complex covariance matrix.
Goodman~\cite{Goodmanb} presents a comprehensive analysis of complex Gaussian models, along with the connection between the class of complex covariance matrices and the Wishart distribution.
Indeed, the complex scaled Wishart distribution is widely adopted as a statistical model for multilook full polarimetric data~\cite{Lopez-MartinezFabregasPottier2005}.

Conradsen \textit{et al}.~\cite{Conradsen2003} proposed a methodology based on the likelihood-ratio test for the discrimination of two Wishart distributed targets, leading to a test statistic that takes into account the complex covariance matrices of PolSAR images.
In a similar fashion, hypothesis tests for monopolarized SAR data were proposed in~\cite{HypothesisTestingSpeckledDataStochasticDistances}.

In this paper, we present analytic expressions for the Kullback-Leibler, R\'enyi (of order $\beta$), Bhattacharyya, and Hellinger distances between scaled complex Wishart distributions in their most general form and in important particular cases.
Frery \textit{et al}~\cite{FreryNascimentoCintraChileanJournalStatistics2011} obtained analytical expressions for these distances, as well as for the $\chi^2$ distance, and they show that the last one is numerically unstable.
Therefore, in the present work tests based on the $\chi^2$ distance were not considered. 

We also verify that those distances present scale invariance with respect to their covariance matrices.    
Using such distances, we derive inequalities which depend on covariance matrices; two among them, obtained from Kullback-Leibler and Hellinger distances, provide alternative forms for deriving the revised Wishart~\cite{ErsahinandCumming2010} and Bartlett~\cite{KerstenandLeeandAinsworth2005} distances, respectively.

Besides advancing the comparison of samples by means of their covariance matrices, the proposed distances are a venue for contrasting images rendered by different numbers of looks.

Considering the hypothesis test methodology proposed by Salicr\'u \textit{et~al.}~\cite{salicruetal1994}, the derived distances are multiplied by a coefficient which involves the sizes of two samples of PolSAR images.
The asymptotic and finite-sample behavior of the resulting quantities is studied.

In order to quantify the sensitivity of the distances, we perform Monte Carlo experiments in several possible scenarios.
We illustrate the behavior of these distances and their associated hypothesis tests with actual data.

This paper unfolds as follows. 
Section~\ref{sec:Wishart} presents the scaled and the relaxed complex Wishart distributions and estimators for their parameters.
Section~\ref{sec:Divergences} recalls the background of stochastic dissimilarities.
Section~\ref{sec:AnalyticResults} presents the analytic expressions of distances between Wishart models, with a new way to derive the Bartlett and the revised Wishart distances.
Section~\ref{sec:Applications} illustrates the application of these distances in PolSAR image discrimination.
Section~\ref{sec:Conclusions} concludes the paper.

\section{The complex Wishart distribution}\label{sec:Wishart}

PolSAR sensors record intensity and relative phase data which can be presented as complex scattering matrices.
In principle, these matrices consist of four complex elements $S_\text{HH},$ $S_\text{HV},$ $S_\text{VH},$ and $S_\text{VV}$, where H and V refer to the horizontal and vertical wave polarization states, respectively.
Under the conditions of the reciprocity theorem~\cite{UlabyElachi1990,LopezMartinezFabregas2003}, we have that $S_\text{HV}=S_\text{VH}$.
This scenario is realistic when natural targets are considered~\cite{Conradsen2003}.

In general, we may consider systems with $p$ polarization elements, which constitute a complex random vector denoted by:
\begin{equation}
\boldsymbol{y}=(S_1\; S_2\;\cdots\; S_p)^{t},
\label{backscattervectorr}
\end{equation}
where the superscript `$t$' indicates vector transposition.
In PolSAR image processing, $\boldsymbol{y}$ is often admitted to obey the multivariate complex circular Gaussian distribution with zero mean~\cite{Goodmanb}, whose probability density function is:
$$
f_{\boldsymbol{y}}({y};\boldsymbol{\Sigma})=\frac{1}{\pi^p|\boldsymbol{\Sigma}|}\exp\bigl(-{y}^{*}\boldsymbol{\Sigma}^{-1}{y}\bigr),
$$
where $|\cdot|$ is the determinant of a matrix or the absolute value of a scalar, the superscript `$*$' denotes the complex conjugate transpose of a vector, $\boldsymbol{\Sigma}$ is the covariance matrix of  $\boldsymbol{y}$ given by
$$
\Sigma=\operatorname{E}(\boldsymbol{y}\boldsymbol{y}^{*}) 
={\left[
\arraycolsep=3.5pt
\begin{array}{cccc}
\operatorname{E}(S_1^{}S_1^{*}) & \operatorname{E}(S_1^{}S_2^{*}) & \cdots &\operatorname{E}(S_1^{}S_p^{*}) \\
\operatorname{E}(S_2^{}S_1^{*}) & \operatorname{E}(S_2^{}S_2^{*}) & \cdots &\operatorname{E}(S_2^{}S_p^{*}) \\
\vdots &  \vdots & \ddots &\vdots \\
\operatorname{E}(S_p^{}S_1^{*}) & \operatorname{E}(S_p^{}S_2^{*}) & \cdots & \operatorname{E}(S_p^{}S_p^{*}) \end{array} \right]},
$$
and $\operatorname{E}\{\cdot\}$ is the statistical expectation operator.
Besides being Hermitian and positive definite, the covariance matrix $\boldsymbol{\Sigma}$ contains all the necessary information to characterize the backscattered data under analysis~\cite{Lopez-MartinezFabregasPottier2005}.

In order to enhance the signal-to-noise ratio, $L$ independent and identically distributed (iid) samples are usually averaged in order to form the $L$-looks covariance matrix~\cite{EstimationEquivalentNumberLooksSAR}:	
$$
\boldsymbol{Z}=\frac{1}{L}\sum_{i=1}^L \boldsymbol{y}_i\boldsymbol{y}_i^{*},
$$
where $\boldsymbol{y}_i$, $i=1,2,\ldots,L$, are realizations of~\eqref{backscattervectorr}.
Under the aforementioned hypotheses, $\boldsymbol{Z}$ follows a scaled complex Wishart distribution. 
Having $\boldsymbol{\Sigma}$ and $L$ as parameters, such law is characterized by the following probability density function:
\begin{equation}
 f_{\boldsymbol{Z}}({Z};\boldsymbol{\Sigma},L) = \frac{L^{pL}|{Z}|^{L-p}}{|\boldsymbol{\Sigma}|^L \Gamma_p(L)} \exp\bigl(
-L\operatorname{tr}\bigl(\boldsymbol{\Sigma}^{-1}{Z}\bigr)\bigr),
\label{eq:denswishart}
\end{equation}
where $\Gamma_p(L)=\pi^{p(p-1)/2}\prod_{i=0}^{p-1}\Gamma(L-i)$, $L\geq p$, $\Gamma(\cdot)$ is the gamma function, and $\operatorname{tr}(\cdot)$ is the trace operator.
This situation is denoted $\boldsymbol{Z}\sim \mathcal W(\boldsymbol{\Sigma},L)$, and this distribution satisfies $\operatorname{E}\{\boldsymbol{Z}\}=\boldsymbol{\Sigma}$, which is a Hermitian positive definite matrix~\cite{EstimationEquivalentNumberLooksSAR}.
In practice, $L$ is treated as a parameter and must be estimated.
In~\cite{AnfinsenJenssenEltoft2009}, Anfinsen~\textit{et~al.} removed the restriction $L\geq p$.
The resulting distribution has the same form as in~\eqref{eq:denswishart} and is termed the relaxed Wishart distribution denoted as $\mathcal{W_R}(\boldsymbol{\Sigma},n)$. 
This model accepts variations of $n$ along the image, and will be assumed henceforth.

Due to its optimal asymptotic properties, the maximum likelihood (ML) estimation is employed to estimate $\boldsymbol{\Sigma}$ and $n$.
Let $\{\boldsymbol{Z}_1, \boldsymbol{Z}_2,\dots, \boldsymbol{Z}_N\}$ be a random sample of size $N$ obeying the $\mathcal{W_R}(\boldsymbol{\Sigma},n)$ distribution. 
If (i)~it is assumed that the parameter $n$ is a known quantity and (ii)~the profile likelihood of $f_{\boldsymbol{Z}}$ is considered in terms of $\boldsymbol{\Sigma}$, we establish the following estimator for $\boldsymbol{\Sigma}$~\cite{Goodmana}:
$$
\widehat{\boldsymbol{\Sigma}}=\frac{1}{N}\sum_{i=1}^N {\boldsymbol{Z}_i}.
$$
Deriving the profile likelihood from~\eqref{eq:denswishart} with respect to $n$ we obtain:
\begin{align} 
\frac{\partial}{\partial n} \ln\bigl[f_{\boldsymbol{Z}}\bigl({Z};&\widehat{\boldsymbol{\Sigma}},n\bigr)\bigr]=p[\log(n)+1]+\log\frac{|{Z}|}{|\widehat{\boldsymbol{\Sigma}}|} \nonumber \\
&- \operatorname{tr}(\widehat{\boldsymbol{\Sigma}}^{-1}{Z})-\sum_{i=0}^{p-1} \psi^{(0)}(n-i),\label{escorefunction}
\end{align}
where $\psi^{(0)}(\cdot)$ is the digamma function~\cite[p. 258]{AbramowitzStegun1994}.
Thus, the solution of above nonlinear equation provides the ML estimator for $n$. 
Several estimation methods for $n$ are discussed in~\cite{EstimationEquivalentNumberLooksSAR}.

\begin{figure}[htb]
\centering
\includegraphics[width=1\linewidth]{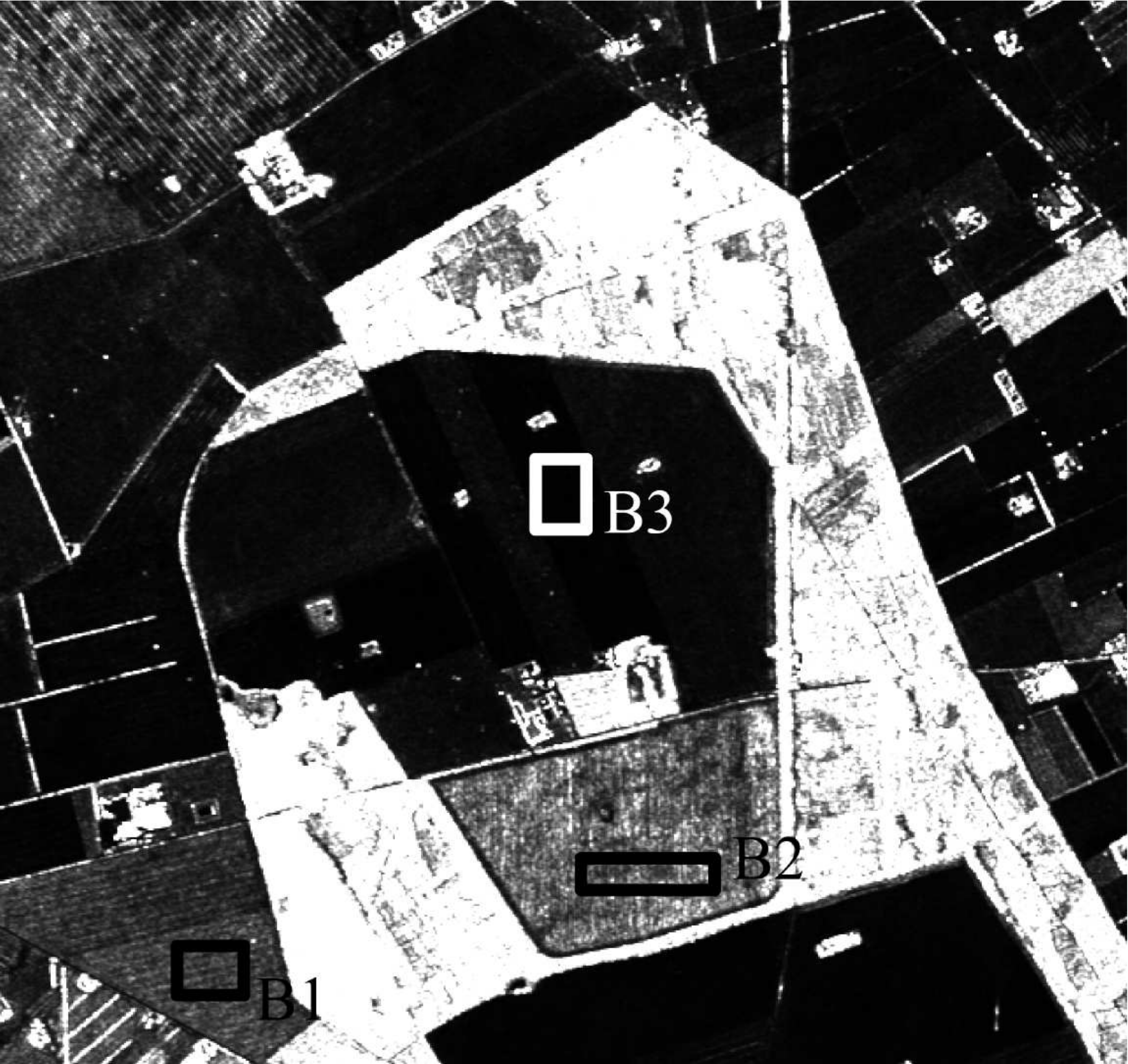}
\caption{EMISAR image (HH channel) with selected regions from Foulum.} 
\label{fig0}
\end{figure}

Fig.~\ref{fig0} presents a polarimetric SAR image obtained by an EMISAR sensor over surroundings of Foulum, Denmark.
The informed (nominal) number of looks is $8$.
According to Skriver \textit{et~al.}~\cite{Skriveretal2005}, the area exhibits three types of crops: (i)~winter rape (B1), (ii)~mixture of winter rape and winter wheat (B2), and (iii)~beets (B3).
Table~\ref{tabelapplica} presents the resulting ML parameter estimates, as well as the sample sizes.
The  closest estimate of $n$ to the nominal number of looks occurs at the most homogeneous scenario, i.e., with beets.
Notice that two out of three ML estimates of the number of looks  are higher than the nominal number of looks.
Similar overestimation was also noticed by Anfinsen \textit{et~al.}~\cite{EstimationEquivalentNumberLooksSAR}, who explained this phenomenon as an effect of the specular reflection on ocean scenarios. 
In our case, winter rape and, to a lesser extent, beets, appear smoother to the sensor than homogeneous targets.

\begin{table}[hbt]                                                  
\centering                                                          
\caption{Parameter estimates on Foulum samples}\label{tabelapplica}                   
\begin{tabular}{c c c c c c } \toprule                       
\multirow{2}{*}{Regions} & \multicolumn{2}{c}{ $\mathcal{W_R}$ }  &  \multirow{2}{*}{\# pixels} \\ \cmidrule(lr{.25em}){2-3} 
 &  \multicolumn{1}{c}{ $\widehat{n}$ } & \multicolumn{1}{c}{$|\widehat{\boldsymbol{\Sigma}}|$}  & \\ \midrule
B1 & 9.216 & 2.507$\times 10^{-6}$  &  1131\\
B2 & 7.200 & 5.717$\times 10^{-6}$  &  1265\\ 
B3 & 8.555 & 4.114$\times 10^{-10}$ &  1155\\
\bottomrule 
%
%
\end{tabular}                                                       
\end{table}                                                         

\begin{figure}[htb]
\centering
\subfigure[B1\label{pasture1}]{\includegraphics[width=.48\linewidth]{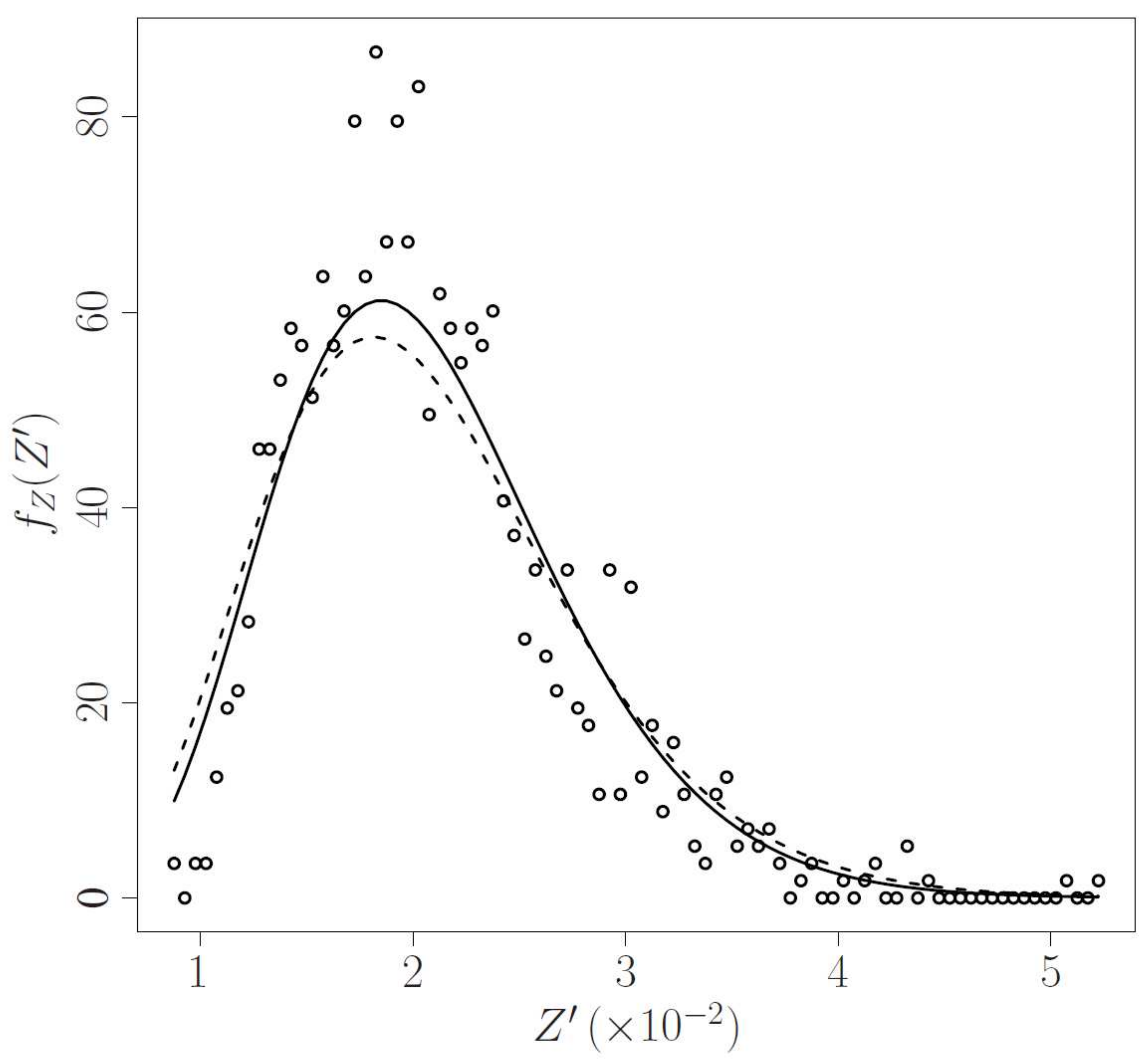}}
\subfigure[B2\label{pasture2}]{\includegraphics[width=.48\linewidth]{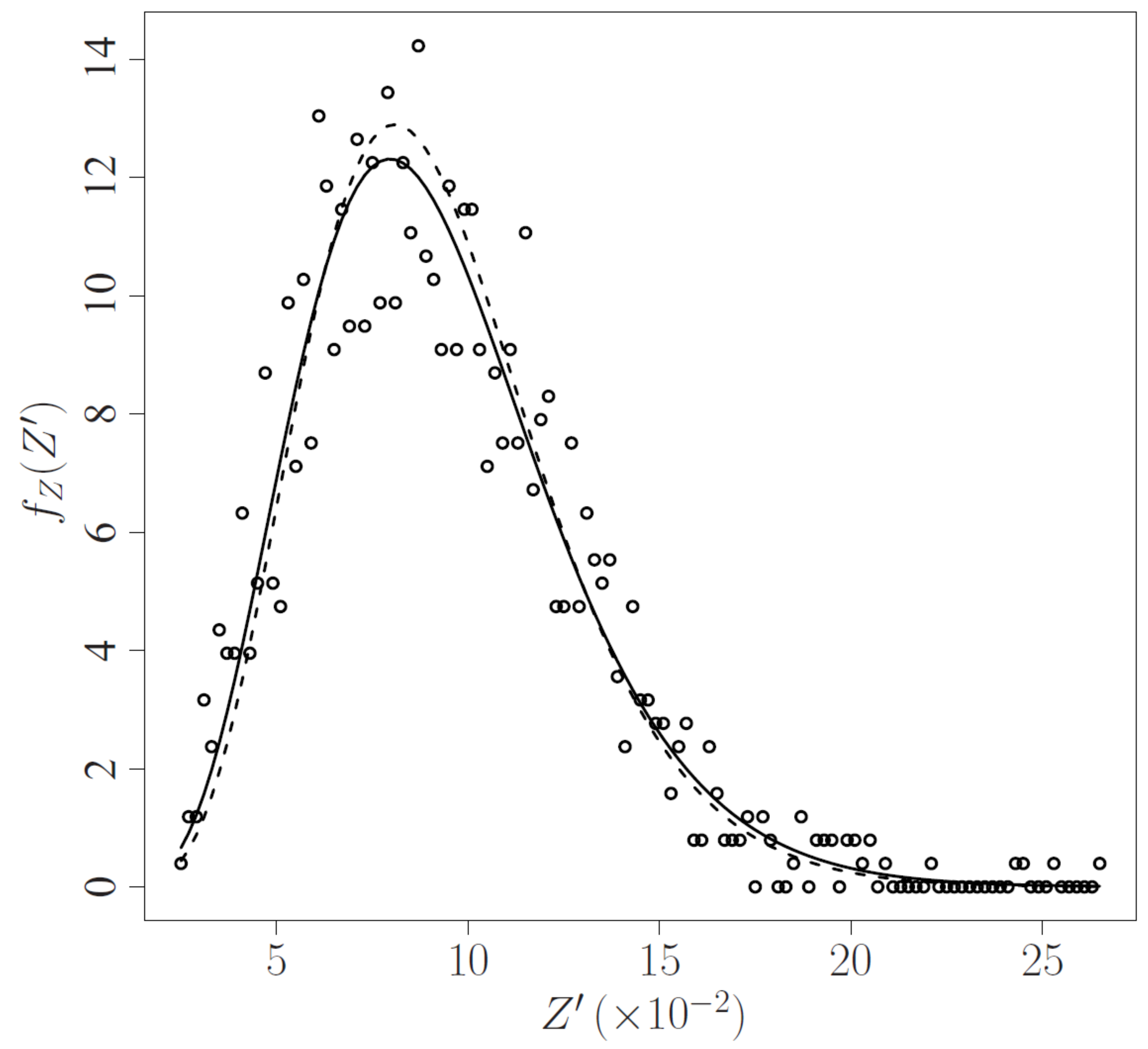}}
\subfigure[B3\label{pasture3}]{\includegraphics[width=.48\linewidth]{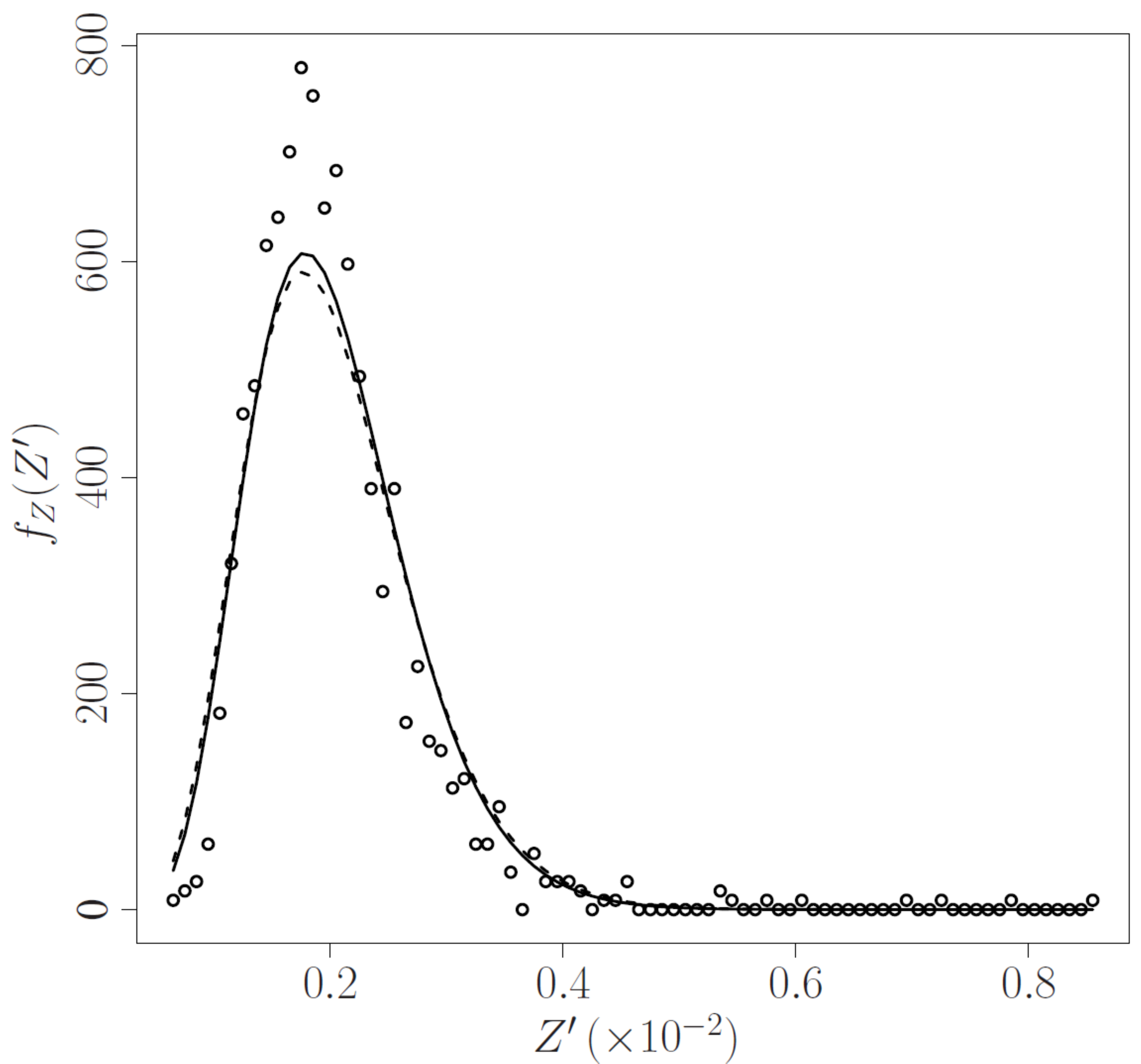}}
\caption{Histograms and empirical relaxed (solid curve) and original (dashed curve) densities of samples.} 
\label{fig1}
\end{figure}

Fig.~\ref{fig1} depicts the empirical densities of data samples from the selected regions.
Additionally, the associated fitted marginal densities $\mathcal{W_R}(\widehat{\boldsymbol{\Sigma}},\widehat{n})$ and $\mathcal{W}(\widehat{\boldsymbol{\Sigma}},8)$  are displayed for comparison.
In this case, the scaled Wishart density collapses to a gamma density as demonstrated in~\cite{Hagedorn2006655}:
$$
f_{Z_i}(Z'_i;n/\sigma^2_{i},n)=\frac{n^n{Z'_i}^{n-1}}{ \sigma^{2n}_i\Gamma(n)}\exp\bigl(-nZ'_i/\sigma^2_i\bigr),
$$
for $i\in\{\text{HH,HV,VV}\}$, where $\sigma^2_{i}$ is the $(i,i)$-th entry of $\boldsymbol{\Sigma}$, and $Z'_i$ is the $(i,i)$-th entry of the random matrix $\boldsymbol{Z}$.  
In order to assess the data fittings, Table~\ref{AICwrelax} presents the Akaike information criterion (AIC) values and the sum of squares due to error (SSE) between the histogram ${f}_k$ of for $Z_{\text{HH}}$, and the fitted densities $\widehat{f}_{Z_{\text{HH}},{\mathcal D}}(Z'_k)$ with ${\mathcal D}\in\{\mathcal{W},\mathcal{W_R}\}$:
$$
\operatorname{SSE}=\sum_{k=1}^{\text{\# pixels}}\frac{(\widehat{f}_{Z_{\text{HH}},{\mathcal D}}(Z'_k)-f_k)^2}{\text{\# pixels}},
$$
where {\# pixels} denote the number of considered pixels.
This measure was used in~\cite{Zhangetal2009}.
In all cases, the $\mathcal{W_R}$ distribution presented the best fit for both measures.
Table~\ref{AICwrelax} also shows the Kolmogorov-Smirnov (KS) statistic and its $p$-value.
It is consistent with the other results, i.e., the scaled Wishart distribution provides better descriptions of the data.

The most accurate fit is in region B2.
The equivalent number of looks in this region is slightly smaller than the nominal one, as expected.
These samples will be used to validate our proposed methods in Section~\ref{sec:InfluenceEstimation}.

\begin{table*}[htb]                                                                                
\centering                                                                                        
\caption{AIC, SSE, and KS statistics values for the HH channel with respect to the relaxed and original Wishart distributions}                        
\label{AICwrelax}
\begin{tabular}{c r@{.}l r@{.}l r@{.}l  r@{.}l r@{.}l r@{.}l }\toprule
\multirow{2}{*}{Regions} &  \multicolumn{4}{c}{AIC} & \multicolumn{4}{c}{SSE} & \multicolumn{4}{c}{KS ($p$-value)}
\\ \cmidrule(lr{.25em}){2-5} \cmidrule(lr{.25em}){6-9} \cmidrule(lr{.25em}){10-13}
& \multicolumn{2}{c}{$\mathcal{W_R}$} & \multicolumn{2}{c}{$\mathcal{W}$} &  \multicolumn{2}{c}{$\mathcal{W_R}$} & \multicolumn{2}{c}{$\mathcal{W}$} & \multicolumn{2}{c}{$\mathcal{W_R}$} & \multicolumn{2}{c}{$\mathcal{W}$}\\ \midrule
B1 &   $-$8401&105 &  $-$8354&817 &   58&169  &   76&853  & 0&070\,(0.325$\times 10^{-4}$) & 0&051\,(0.006) \\
B2 &   $-$4973&743 &  $-$4987&579 &    1&245  &    1&487  & 0&018\,(0.789)   & 0&034\,(0.110) \\
B3 &  $-$13725&650 & $-$13734&330 & 2362&108  & 2836&599  & 0&063\,(2.383$\times 10^{-4}$) & 0&059\,(6.435$\times 10^{-4}$)\\
\bottomrule                                                                                       
\end{tabular}                                                                                     
\end{table*}                   

\section{Stochastic dissimilarities}\label{sec:Divergences}

In the following we adhere to the convention that a ``divergence'' is any non-negative function between two probability measures which obeys the identity of definiteness property~\cite[ch. 11, p. 328]{BartleandSherbert1982}.
If the function is also symmetric, it is called a ``distance''.
Finally, we understand ``metric'' as a distance which also satisfies the triangular inequality~\cite[ch. 1 and 14]{deza2009encyclopedia}.

An image can be understood as a set of regions, in which the enclosed pixels are observations of random variables following a certain distribution.
Therefore, stochastic dissimilarity measures can be used as image processing tools, since they may be able to assess the difference between the distributions that describe different image areas~\cite{HypothesisTestingSpeckledDataStochasticDistances}.
Dissimilarity measures were submitted to a systematic and comprehensive treatment in \cite{Ali1996,Csiszar1967} and, as a result, the class of $(h,\phi)$-divergences was proposed \cite{salicruetal1994}.

Assume that $\boldsymbol{X}$ and $\boldsymbol{Y}$ are random matrices associated with densities $f_{\boldsymbol{X}}(Z;\boldsymbol{\theta_1})$ and $f_{\boldsymbol{Y}}(Z;\boldsymbol{\theta_2})$, respectively, where $\boldsymbol{\theta_1}$ and $\boldsymbol{\theta_2}$ are parameter vectors.
The densities are assumed to share a common support $\boldsymbol{\mathcal A}$: the cone of Hermitian positive definite matrices~\cite{AnfinsenDoulgerisEltoft2011}.
The $(h,\phi)$-divergence between $f_{\boldsymbol{X}}$ and $f_{\boldsymbol{Y}}$ is defined by
\begin{equation} 
D_{\phi}^h(\boldsymbol{X},\boldsymbol{Y}) = 
h\biggl(\int_{\boldsymbol{\mathcal A}} \phi\biggl( \frac{f_{\boldsymbol{X}}({Z};\boldsymbol{\theta_1})}{f_{\boldsymbol{Y}}({Z};\boldsymbol{\theta_2})}\biggr) f_{\boldsymbol{Y}}({Z};\boldsymbol{\theta_2})\mathrm{d}{Z}\biggr),
\label{eq:eps2-no}
\end{equation}
where $h\colon(0,\infty)\rightarrow[0,\infty)$ is a strictly increasing function with $h(0)=0$, $\phi\colon (0,\infty)\rightarrow[0,\infty)$ is a convex function, and indeterminate forms are assigned the value zero (we assume the conventions (i)~$\phi(0)=\lim_{x \downarrow 0}\,f(x)$, (ii)~$0\,\phi(0/0) \equiv 0$, and, for $a>0$, (iii)~$0\,\phi(a/0)=\lim_{\epsilon \downarrow 0} \,\epsilon\,\phi(a/\epsilon)=a\,\lim_{x\rightarrow \infty}\,\phi(x)/x$) \cite[pp.~31]{csiszar2004information}.
In particular, Ali and Silvey~\cite{Ali1996} proposed a detailed discussion about the function $\phi$.
 The differential element $\mathrm{d}{Z}$ is given by
$$
\mathrm{d}{Z}=\mathrm{d}Z_{11}\mathrm{d}Z_{22}\cdots\mathrm{d}Z_{pp}\displaystyle\prod^p_{\underbrace{i,j=1}_{i<j}}\mathrm{d}\Re\{Z_{ij}\} \mathrm{d}\Im\{Z_{ij}\},
$$ 
where $Z_{ij}$ is the $(i,i)$-th entry of matrix ${Z}$, and operators $\Re\{\cdot\}$ and $\Im \{\cdot\}$ return real and imaginary parts of their arguments, respectively~\cite{Goodmanb}. 

Well-known divergences arise after adequate choices of $h$ and $\phi$. 
Among them, we examined the following: (i)~Kullback-Leibler~\cite{SeghouaneAmari2007}, (ii)~R\'enyi, (iii)~Bhattacharyya~\cite{Kailath1967}, and (iv)~Hellinger~\cite{HypothesisTestingSpeckledDataStochasticDistances}.
As the triangular inequality is not necessarily satisfied, not every divergence measure is a metric~\cite{BurbeaRao1982}.
Additionally, the symmetry property is not followed by some of these divergence measures.
Nevertheless, such tools are mathematically appropriate for comparing the distribution of random variables~\cite{JagerWellner2007}.
The following expression has been suggested as a possible solution for this issue~\cite{SeghouaneAmari2007}:
$$
d_{\phi}^h(\boldsymbol{X},\boldsymbol{Y})=\frac{D_{\phi}^h(\boldsymbol{X},\boldsymbol{Y})+D_{\phi}^h(\boldsymbol{Y},\boldsymbol{X})}{2}.
$$

Functions $d_{\phi}^h:\boldsymbol{\mathcal A} \times \boldsymbol{\mathcal A} \rightarrow \mathbb{R}$ are distances over $\boldsymbol{\mathcal A}$ since, for all $\boldsymbol{X},\boldsymbol{Y}\in \boldsymbol{\mathcal A}$, the following properties hold:
\begin{enumerate}
\item $d_{\phi}^h(\boldsymbol{X},\boldsymbol{Y})\geq 0$ (Non-negativity).
\item $d_{\phi}^h(\boldsymbol{X},\boldsymbol{Y})=d_{\phi}^h(\boldsymbol{Y},\boldsymbol{X})$ (Symmetry).
\item $d_{\phi}^h(\boldsymbol{X},\boldsymbol{Y})=0 \Leftrightarrow \boldsymbol{X}=\boldsymbol{Y} $ (Definiteness).
\end{enumerate}
Table~\ref{tab-1} shows the functions $h$ and $\phi$ which lead to the distances considered in this work.

\begin{table*}[hbt]
\centering   
\caption{($h,\phi$)-distances and their functions}
\begin{tabular}{ccc}
\toprule
{ $(h,\phi)$-{distance}} & { $h(y)$} & { $\phi(x)$} \\
\midrule
Kullback-Leibler & ${y}/{2}$ & $(x-1)\log x$  \\
R\'{e}nyi (order $\beta$) & $\frac{1}{\beta-1}\log((\beta-1)y+1),\;0\leq y<\frac{1}{1-\beta}$ & 
$\frac{x^{1-\beta}+x^{\beta}-\beta(x-1)-2}{2(\beta-1)},0<\beta<1$\\
Bhattacharyya  & $-\log(-y+1),0\leq y<1$ & $-\sqrt{x}+\frac{x+1}{2}$ \\
Hellinger  & ${y}/{2},0\leq y<2$ &  $\,\,\,(\sqrt{x}-1)^2$  \\ \bottomrule
\end{tabular}
\label{tab-1}
\end{table*}

In the following we discuss integral expressions of these $(h,\phi)$-distances. 
For simplicity, we suppress the explicit dependence on $Z$ and $(\boldsymbol{\theta}_1,\boldsymbol{\theta}_2)$, reminding that the integration is with respect to $Z$ on $\boldsymbol{\mathcal A}$.
\begin{itemize}
\item [(i)] The Kullback-Leibler distance: 
\begin{align*} 
d_{\text{KL}}(\boldsymbol{X},\boldsymbol{Y})&=\frac 12 [D_{\text{KL}}(\boldsymbol{X},\boldsymbol{Y})+D_{\text{KL}}(\boldsymbol{Y},\boldsymbol{X})]\\
&= \frac 12 \biggl[ \int f_{\boldsymbol{X}}\log{\frac{f_{\boldsymbol{X}}}{f_{\boldsymbol{Y}}}} + \int f_{\boldsymbol{Y}}\log{\frac{f_{\boldsymbol{Y}}}{f_{\boldsymbol{X}}}} \biggl] \\
&=\frac{1}{2}\int(f_{\boldsymbol{X}}-f_{\boldsymbol{Y}})\log{\frac{f_{\boldsymbol{X}}}{f_{\boldsymbol{Y}}}}.
\end{align*}
The divergence $D_{\text{KL}}$ has a close relationship with the Neyman-Pearson lemma~\cite{est3} and its symmetrization has been suggested as a correction form of the Akaike information criterion~\cite{SeghouaneAmari2007}.
\item [(ii)] The R\'{e}nyi distance of order $\beta$:
\begin{align*} 
\widetilde{d_{\text{R}}^{\beta }}(\boldsymbol{X},\boldsymbol{Y})&= \frac 12 [ D_\text{R}^{\beta}(\boldsymbol{X},\boldsymbol{Y})+D_\text{R}^{\beta}(\boldsymbol{	Y},\boldsymbol{X})]\\ 
&=\frac{ \log\int f_{\boldsymbol{X}}^{\beta}f_{\boldsymbol{Y}}^{1-\beta} + \log \int f_{\boldsymbol{X}}^{1-\beta}f_{\boldsymbol{Y}}^{\beta}}{2(\beta-1)},
\end{align*}
where $0<\beta<1$.
The divergence $D_\text{R}^{\beta}$ has been used for analysing geometric characteristics with respect to probability laws~\cite{Andai2009777}.
By the Fej\'er inequality~\cite{Neuman1990}, we have that 
\begin{align*}
d_{\text{R}}^{\beta}(\boldsymbol{X},\boldsymbol{Y}) \triangleq & \frac{1}{\beta-1}\log   \frac{ \int f_{\boldsymbol{X}}^{\beta}f_{\boldsymbol{Y}}^{1-\beta} + \int f_{\boldsymbol{X}}^{1-\beta}f_{\boldsymbol{Y}}^{\beta} }{2}\\
& \leq \widetilde{d_{\text{R}}^{\beta }}(\boldsymbol{X},\boldsymbol{Y}).
\end{align*}
The distance ${d_{\text{R}}^{\beta}}$ proves to be more algebraically tractable than $\widetilde{d_{\text{R}}^{\beta}}$ for some manipulations with the complex Wishart density.
Thus, we use the former in subsequent analyses. 

\item [(iii)] The Bhattacharyya distance:
$$
d_{\text{B}}(\boldsymbol{X},\boldsymbol{Y})= -\log \int\sqrt{f_{\boldsymbol{X}}f_{\boldsymbol{Y}}}.
$$
Goudail~\textit{et~al.}~\cite{GoudailRefregierDelyon2004} showed that this distance is an efficient tool for contrast definition in algorithms for image processing.

\item [(iv)] The Hellinger distance:
$$
d_{\text{H}}(\boldsymbol{X},\boldsymbol{Y})=1-\int\sqrt{f_{\boldsymbol{X}}f_{\boldsymbol{Y}}}.
$$
Estimation methods based on the minimization of $d_{\text{H}}$ have been successfully employed in the context of stochastic differential equations~\cite{Gietandmichel2008}.  
This is the only bounded distance among the ones considered in this paper.
\end{itemize}

When considering the distance between particular cases of the same distribution, only parameters are relevant.
In this case, the parameters $\boldsymbol{\theta_1}$ and $\boldsymbol{\theta_2}$ replace the random variables $\boldsymbol{X}$ and $\boldsymbol{Y}$ as arguments of the discussed distances.
This notation is in agreement with that of \cite{salicruetal1994}.

In the following, the hypothesis test based on stochastic distances proposed by Salicr\'u \textit{et~al.}~\cite{salicruetal1994} is introduced.   
Let $M$-point vectors $\widehat{\boldsymbol{\theta}}_1=(\widehat{\theta}_{11},\ldots,\widehat{\theta}_{1M})$ and $\widehat{\boldsymbol{\theta}}_2=(\widehat{\theta}_{21},\ldots,\widehat{\theta}_{2M})$ be the ML estimators of parameters $\boldsymbol{\theta_1}$ and $\boldsymbol{\theta_2}$ based on independent samples of sizes $N_1$ and $N_2$, respectively.
Under the regularity conditions discussed in \cite[p. 380]{salicruetal1994} the following lemma holds:
\begin{Lemma}\label{prop-chi}
If $\frac{N_1}{N_1+N_2} \xrightarrow[N_1,N_2\rightarrow\infty]{} \lambda\in(0,1)$ and $\boldsymbol{\theta}_1=\boldsymbol{\theta}_2$, then
\begin{equation}
S_{\phi}^h(\widehat{\boldsymbol{\theta}}_1,\widehat{\boldsymbol{\theta}}_2)=\frac{2 N_1 N_2}{N_1+N_2}\frac{d^h_{\phi}(\widehat{\boldsymbol{\theta}}_1,\widehat{\boldsymbol{\theta}}_2)}{ h{'}(0) \phi{''}(1)}   \xrightarrow[N_1,N_2\rightarrow\infty]{\mathcal D}\chi_{M}^2,
\label{eq:chi2stat1}
\end{equation}
where ``$\xrightarrow[]{\mathcal{D}}$'' denotes convergence in distribution and $\chi_{M}^2$ represents the chi-square distribution with $M$ degrees of freedom. 
\end{Lemma}

Based on Lemma~\ref{prop-chi}, statistical hypothesis tests for the null hypothesis $\boldsymbol{\theta}_1=\boldsymbol{\theta}_2$ can be derived in the form of the following proposition.
\begin{Proposition}
Let $N_1$ and $N_2$ be large and $S_{\phi}^h(\widehat{\boldsymbol{\theta}}_1,\widehat{\boldsymbol{\theta}}_2)=s$, then the null hypothesis $\boldsymbol{\theta}_1=\boldsymbol{\theta}_2$ can be re\-jec\-ted at le\-vel $\alpha$ if $\Pr( \chi^2_{M}>s)\leq \alpha$. 
\label{p-3}
\end{Proposition}

We denote the statistics based on the Kullback-Leibler, R\'enyi, Bhattacharya, and Hellinger distances as $S_\text{KL}$, $S_\text{R}^{\beta}$, $S_\text{B}$, and $S_\text{H}$, respectively.

\section{Analytic expressions, sensitivity, inequalities, and finite sample size behavior}\label{sec:AnalyticResults}

In the following,  analytic expressions for the stochastic distances $d_{\text{KL}}$, $d_\text{R}^\beta$, $d_{\text{B}}$, and $d_{\text{H}}$ between two relaxed complex Wishart distributions are derived (Section~\ref{sec:fullform}).
%
We examine the special cases in terms of the parameter values: (i)~$\boldsymbol{\Sigma}_1\neq\boldsymbol{\Sigma}_2$ and $n_1\neq n_2$, which correspond to the most general case, (ii)~same equivalent number of looks $n_1=n_2=n$ and different covariance matrices $\boldsymbol{\Sigma}_1\neq\boldsymbol{\Sigma}_2$, and (iii)~same covariance matrix $\boldsymbol{\Sigma}_1=\boldsymbol{\Sigma}_2$ and different equivalent number of looks $n_1\neq n_2$. 
Case~(ii) is likely to be the most frequently used in practice since it allows the comparison of two possibly different areas from the same image.
Case~(iii) allows the assessment of a change in distribution due only to multilook processing on the same area.

The sensitivity of the tests to variations of parameters is qualitatively assessed and discussed in Section~\ref{sec:plots}.

In Section~\ref{sec:inequality} we derive inequalities which $\boldsymbol{\Sigma}_1$ and $\boldsymbol{\Sigma}_2$ must obey.
These inequalities lead to the Bartlett and revised Wishart distances in a different and simple way when compared to a well-known method available in literature~\cite{leeetal1994a}.
Distances are also shown to satisfy scale invariance with respect to $\boldsymbol{\Sigma}$.


The performance of the tests for finite size samples is quantified by means of (i)~Monte Carlo simulation and (ii)~true data analysis in Section~\ref{sec:InfluenceEstimation}.

\subsection{Analytic expressions}\label{sec:fullform}

\subsubsection{Kullback-Leibler distance} 

\begin{description}
\item[Case (i):] 
\begin{align}
&d_\text{KL}(\boldsymbol{\theta}_1 ,\boldsymbol{\theta}_2)=\frac{n_1-n_2}{2}\bigg\{\log\frac{|\boldsymbol{\Sigma}_1|}{|\boldsymbol{\Sigma}_2|}-p\log\frac{n_1}{n_2}\nonumber \\
&+p\bigl[\psi^{(0)}(n_1-p+1)-\psi^{(0)}(n_2-p+1)\bigr]\nonumber \\
&+(n_2-n_1)\displaystyle \sum_{i=1}^{p-1}\frac{i}{(n_1-i)(n_2-i)}\bigg\} \nonumber \\
&+ \frac{\operatorname{tr}(n_2\boldsymbol{\Sigma}_2^{-1}\boldsymbol{\Sigma}_1+n_1\boldsymbol{\Sigma}_1^{-1}\boldsymbol{\Sigma}_2)}{2}-\frac{p(n_1+n_2)}{2}.
\label{expreKL}
\end{align}
Details of this derivation are given in Appendix~\ref{app:KL}.

\item [Case (ii):] 
$$ 
d_\text{KL}(\boldsymbol{\theta}_1,\boldsymbol{\theta}_2)=n\bigg[\frac{\operatorname{tr}(\boldsymbol{\Sigma}_1^{-1}\boldsymbol{\Sigma}_2+\boldsymbol{\Sigma}_2^{-1}\boldsymbol{\Sigma}_1)}{2}-p\bigg].  
$$
This result was also derived by Lee and Bretschneider~\cite{LeeandBretschneider2011} and applied to real PolSAR data for assessing separability of target classes.

\item [Case (iii):] 
\begin{align*}
&d_\text{KL}(\boldsymbol{\theta}_1,\boldsymbol{\theta}_2)=\frac{n_1-n_2}{2}\bigg\{-p\log\frac{n_1}{n_2}\\
&+p\bigl[\psi^{(0)}(n_1-p+1)-\psi^{(0)}(n_2-p+1)\bigr]\\
&+(n_2-n_1)\displaystyle \sum_{i=1}^{p-1}\frac{i}{(n_1-i)(n_2-i)}\bigg\}.
\end{align*}
\end{description}

\subsubsection{R\'enyi distance of order $0<\beta<1$}

\begin{description}
\item[Case (i):] 
\begin{equation}
d_\text{R}^{\beta}(\boldsymbol{\theta}_1,\boldsymbol{\theta}_2)=\frac{1}{\beta-1}\log  \frac{I(\boldsymbol{\theta}_1,\boldsymbol{\theta}_2)}{2},
\label{distREnyi}
\end{equation}
where 
\begin{align*}
&I(\boldsymbol{\theta}_1,\boldsymbol{\theta}_2)=\\
&\\
&\biggl[\frac{\Gamma(n_1-p+1)^p}{n_1^{pn_1}} |\boldsymbol{\Sigma}_1|^{n_1} \displaystyle \prod_{i=1}^{p-1}(n_1-i)^i \biggr]^{-\beta}\\
&\\
\times&\biggl[\frac{\Gamma(n_2-p+1)^p}{n_2^{pn_2}} |\boldsymbol{\Sigma}_2|^{n_2} \displaystyle \prod_{i=1}^{p-1}(n_2-i)^i \biggr]^{\beta-1}\\
&\\
\times&\,\,\,\Gamma(E_{12}-p+1)^p  |\boldsymbol{\Sigma}_{12}|^{E_{12}} \displaystyle \prod_{i=1}^{p-1}(E_{12}-i)^i\\ 
&\\
+&\Bigl[\frac{\Gamma(n_1-p+1)^p}{n_1^{pn_1}} |\boldsymbol{\Sigma}_1|^{n_1} \displaystyle \prod_{i=1}^{p-1}(n_1-i)^i \biggr]^{\beta-1}\\
&\\
\times&\biggl[\frac{\Gamma(n_2-p+1)^p}{n_2^{pn_2}} |\boldsymbol{\Sigma}_2|^{n_2} \displaystyle \prod_{i=1}^{p-1}(n_2-i)^i \biggr]^{-\beta}\\
&\\
\times&\,\,\,\Gamma(E_{21}-p+1)^p  |\boldsymbol{\Sigma}_{21}|^{E_{21}} \displaystyle \prod_{i=1}^{p-1}(E_{21}-i)^i,\\
\end{align*}
where $E_{ij}=\beta n_i+(1-\beta)n_j$, 
for $i,j=1,2$, and $
\boldsymbol{\Sigma}_{ij}=|(n_i\beta \boldsymbol{\Sigma}_i^{-1}+n_j(1-\beta)\boldsymbol{\Sigma}_j^{-1})^{-1}|$.

\item [Case (ii):] 
\begin{align*}
d_\text{R}^{\beta}(\boldsymbol{\theta}_1,\boldsymbol{\theta}_2)=&\frac{\log 2}{1-\beta}+\frac{1}{\beta-1}\log\Big\{ \\
&\Bigl[\frac{|(\beta \boldsymbol{\Sigma}_1^{-1}+(1-\beta)\boldsymbol{\Sigma}_2^{-1})^{-1}|}{|\boldsymbol{\Sigma}_1|^{\beta}|\boldsymbol{\Sigma}_2|^{1-\beta}}\Bigr]^n \\
+&\Bigl[\frac{|(\beta \boldsymbol{\Sigma}_2^{-1}+(1-\beta)\boldsymbol{\Sigma}_1^{-1})^{-1}|}{|\boldsymbol{\Sigma}_1|^{(1-\beta)}|\boldsymbol{\Sigma}_2|^{\beta}}\Bigr]^n\Big\}.
\end{align*}

\item [Case (iii):] 
\begin{align*}
&d_\text{R}^{\beta}(\boldsymbol{\theta}_1,\boldsymbol{\theta}_2)=\frac{\log 2}{1-\beta}+\frac{1}{\beta-1}\log\biggl\{ \\
&\biggl[\Gamma(n_1-p+1)^p n_1^{-pn_1} \displaystyle \prod_{i=1}^{p-1}(n_1-i)^i \biggr]^{-\beta}\\
\times&\biggl[\Gamma(n_2-p+1)^p n_2^{-pn_2} \displaystyle \prod_{i=1}^{p-1}(n_2-i)^i \biggr]^{\beta-1} \\
\times&\Gamma(E_1-p+1)^p  {E_1}^{-pE_1} \displaystyle \prod_{i=1}^{p-1}(E_1-i)^i \\
+&\Bigl[\Gamma(n_1-p+1)^p n_1^{-pn_1} \displaystyle \prod_{i=1}^{p-1}(n_1-i)^i \biggr]^{\beta-1}\\
\times&\biggl[\Gamma(n_2-p+1)^p n_2^{-pn_2} \displaystyle \prod_{i=1}^{p-1}(n_2-i)^i \biggr]^{-\beta} \\
\times&\Gamma(E_2-p+1)^p  {E_2}^{-pE_2} \displaystyle \prod_{i=1}^{p-1}(E_2-i)^i \biggr\}.
\end{align*}
\end{description}

\subsubsection{Bhattacharyya distance}

\begin{description} 
\item[Case (i):] 
\begin{align}\label{expreBA}
&d_\text{B}(\boldsymbol{\theta}_1,\boldsymbol{\theta}_2)=\frac{n_1\log|\boldsymbol{\Sigma}_1|}{2}+\frac{n_2\log|\boldsymbol{\Sigma}_2|}{2}\nonumber \\
&\mbox{}-\frac{n_1+n_2}{2}\log\bigg|\bigg(\frac{n_1\boldsymbol{\Sigma}_1^{-1}+n_2\boldsymbol{\Sigma}_2^{-1}}{2}\bigg)^{-1} \bigg|\nonumber \\
&\mbox{}+\sum_{k=0}^{p-1}\log\frac{\sqrt{\Gamma(n_1-k)\Gamma(n_2-k)}}{\Gamma(\frac{n_1+n_2}{2}-k)}\nonumber \\
&\mbox{}-\frac{p}{2}(n_1\log{n_1}+n_2\log{n_2}).
\end{align}

\item[Case (ii):]
\begin{align*}
d_\text{B}(\boldsymbol{\theta}_1,\boldsymbol{\theta}_2)=&n\bigg[\frac{\log|\boldsymbol{\Sigma}_1|+\log|\boldsymbol{\Sigma}_2|}{2} \\
&\mbox{}-\log\bigg|\bigg(\frac{\boldsymbol{\Sigma}_1^{-1}+\boldsymbol{\Sigma}_2^{-1}}{2}\bigg)^{-1}\bigg|\bigg].
\end{align*}

\item[Case (iii):]
\begin{align*}
&d_\text{B}(\boldsymbol{\theta}_1,\boldsymbol{\theta}_2)=p\frac{n_1+n_2}{2}\log\frac{n_1+n_2}{2}\\
&\mbox{}+\sum_{k=0}^{p-1}\log\frac{\sqrt{\Gamma(n_1-k)\Gamma(n_2-k)}}{\Gamma(\frac{n_1+n_2}{2}-k)}\\
&\mbox{}-\frac{p}{2}(n_1\log{n_1}+n_2\log{n_2}).
\end{align*}
\end{description}

\subsubsection{Hellinger distance}

\begin{description}
\item[Case (i):] 
\begin{align}
 d_\text{H}(\boldsymbol{\theta}_1,&\boldsymbol{\theta}_2) = 1- \sqrt{n_1^{pn_1}n_2^{pn_2}} \nonumber\\
&\times\frac{\bigl|2^{-1}({n_1\boldsymbol{\Sigma}_1^{-1}+n_2\boldsymbol{\Sigma}_2^{-1}})^{-1}\bigr|^{\frac{n_1+n_2}{2}}}{|\boldsymbol{\Sigma}_1|^{\frac{n_1}{2}}|\boldsymbol{\Sigma}_2|^{\frac{n_2}{2}}}\nonumber\\
&\times \prod_{k=0}^{p-1}\frac{\Gamma\bigl(\frac{n_1+n_2}{2}-k\bigr)}{\sqrt{\Gamma({n_1}{}-k)\Gamma({n_2}{}-k)}}.
\label{distHELn}
\end{align}

\item[Case (ii):] 
$$
\hskip-5em
d_\text{H}(\boldsymbol{\theta}_1,\boldsymbol{\theta}_2) =1-\Bigg[\frac{\bigl|2^{-1}({\boldsymbol{\Sigma}_1^{-1}+\boldsymbol{\Sigma}_2^{-1}})^{-1}\bigr|}{\sqrt{|\boldsymbol{\Sigma}_1||\boldsymbol{\Sigma}_2|}}\Bigg]^n. 
$$

\item[Case (iii):]
\begin{align*}
\hskip-5em
d_\text{H}(\boldsymbol{\theta}_1,&\boldsymbol{\theta}_2)=1-\sqrt{n_1^{pn_1}n_2^{pn_2}}\\
&\times {\Big(\frac{n_1+n_2}{2}\Big)^{-p\frac{n_1+n_2}{2}}}\\
&\times  \prod_{k=0}^{p-1}\frac{\Gamma\big(\frac{n_1+n_2}{2}-k\big)}{\sqrt{\Gamma({n_1}{}-k)\Gamma({n_2}{}-k)}}.
\end{align*}
\end{description}

\subsection{Sensitivity analysis} \label{sec:plots}


Now we examine the behavior of the statistics presented in Lemma~\ref{prop-chi} with respect to parameter variations, i.e., under the  alternative hypotheses.
These statistics are directly comparable since they all have the same asymptotic distribution; we used $N_1=N_2=100$.
Two simple alternative hypotheses are illustrated: changes in an entry in the diagonal of the covariance matrix, and changes in the number of looks.

Firstly, we assumed $n=8$ looks, $\boldsymbol{\theta}_1=(\boldsymbol{\Sigma}(360932),8)$ and $\boldsymbol{\theta}_2=(\boldsymbol{\Sigma}(x),8)$, where
\begin{equation}
\boldsymbol{\Sigma}(x)={\left[\begin{array}{ccc} x & 11050+3759\textbf{i} & 63896+1581\textbf{i} \\ 
                & 98960 & 6593+6868\textbf{i} \\
                &  & 208843  \end{array} \right]}.
\label{matrixEX}								
\end{equation}
Since the covariance matrix is Hermitian, only the upper triangle and the diagonal are displayed.
The fixed covariance matrix $\boldsymbol{\Sigma}(360932)$ was previously analyzed in~\cite{PolarimetricSegmentationBSplinesMSSP} in PolSAR data of forested areas.

\begin{figure}[htb] 
\centering
\subfigure[Varying $\boldsymbol{\Sigma} $\label{CurvesWishart2}]{\includegraphics[width=.48\linewidth]{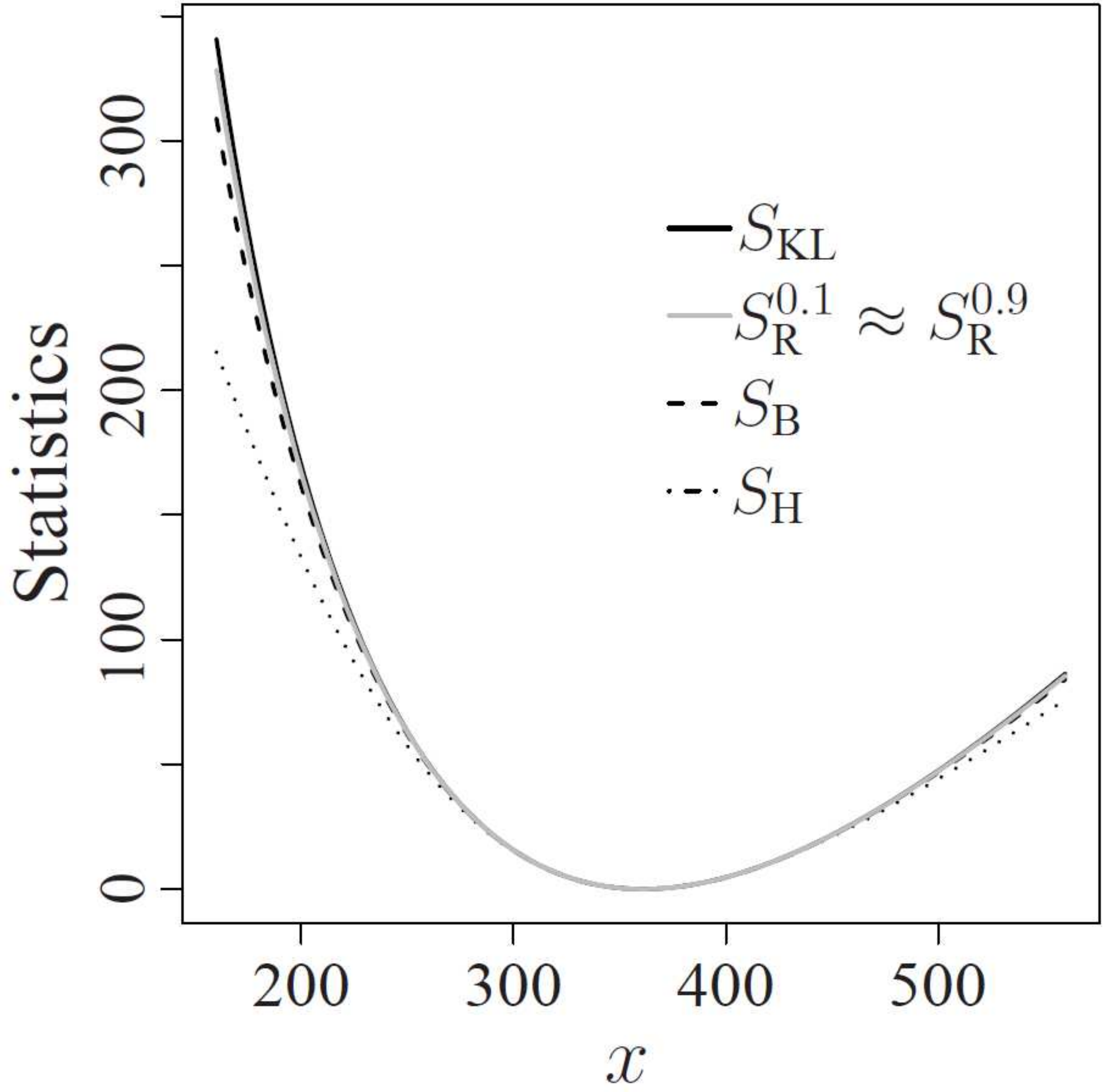}}
\subfigure[Varying $n$ \label{CurvesWishart1}]{\includegraphics[width=.48\linewidth]{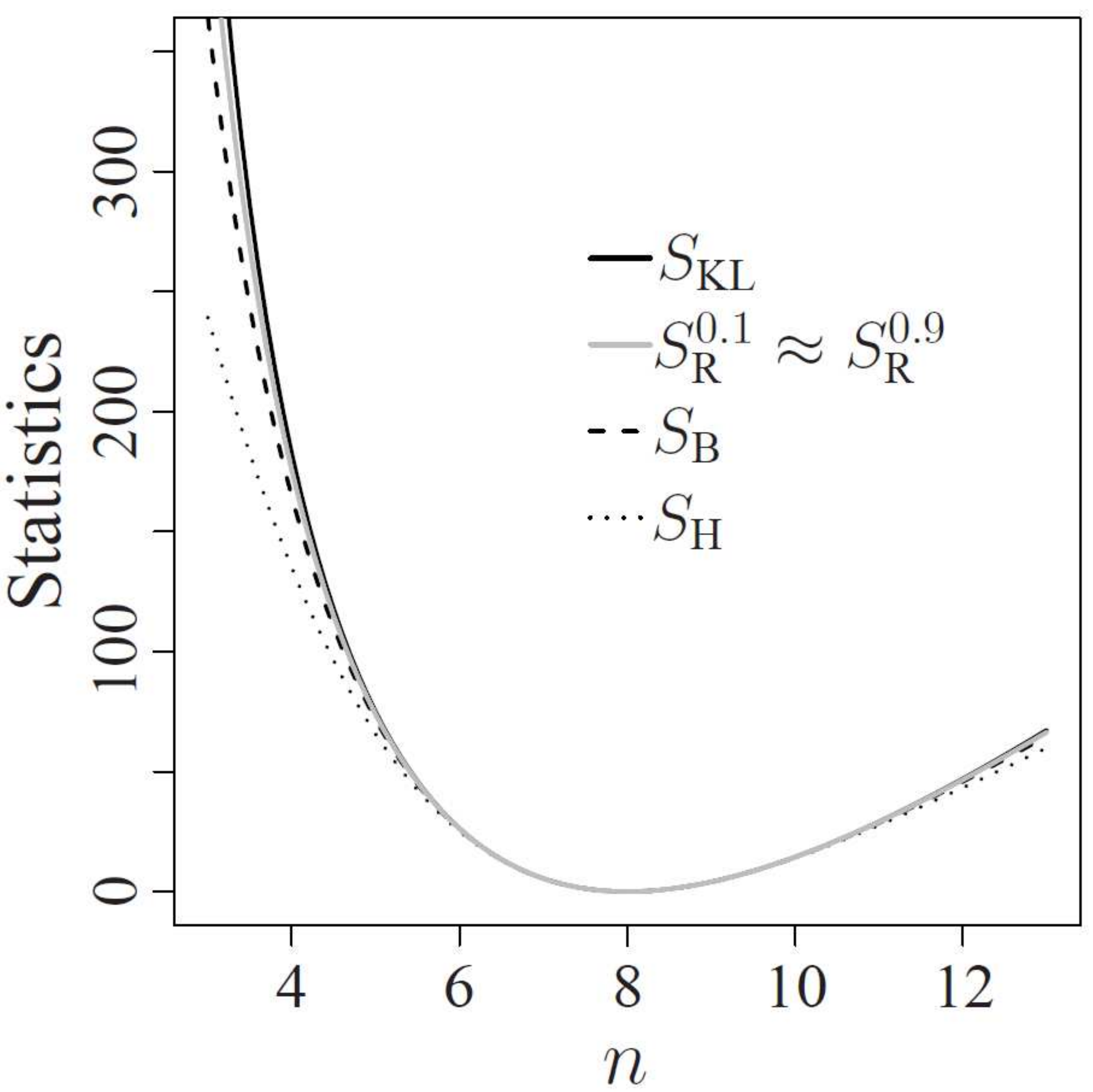}}
\caption{Sensitivity of statistics.} 
\label{CurvesWishart}
\end{figure}

Fig.~\ref{CurvesWishart2} shows the statistics for $x \in [160000,560000]$.
They present roughly the same behavior.

Secondly, we considered fixed covariance matrices with varying equivalent number of looks: $\boldsymbol{\theta}_1=(\boldsymbol{\Sigma}(360932),8)$ and $\boldsymbol{\theta}_2=(\boldsymbol{\Sigma}(360932),m)$, for $3\leq m \leq 13$.
Fig.~\ref{CurvesWishart1} shows the statistics.
It is noticeable that the test statistics are steeper to the left of the minimum.
The number of looks, being a shape parameter, alters the distribution in a nonlinear fashion. Such change is perceived visually and by distance measures, and it is more intense for low values of the parameter.
In other words, the difference between $\mathcal{W_R}(\mathbf\Sigma,n)$ and $\mathcal{W_R}(\mathbf\Sigma,kn)$, for any fixed $k>1$ and any $\mathbf\Sigma$, becomes smaller when $n$ increases. 
            
\subsection{Invariance and inequalities}\label{sec:inequality}

The derived distances are invariant under scalings of the covariance matrix $\boldsymbol{\Sigma}$.
In fact, it can be shown that
$$
d_{\mathcal M}[(a\boldsymbol{\Sigma}_1,n_1),(a\boldsymbol{\Sigma}_2,n_2)]= d_{\mathcal M}[(\boldsymbol{\Sigma}_1,n_1),(\boldsymbol{\Sigma}_2,n_2)], 
$$
where $a$ is a positive real value and  $\mathcal M \in \{\text{KL},\text{R},\text{B},\text{H}\}$.
This fact stems directly from the mathematical definition of these distances.

{De Maio} and Alfano~\cite{AntonioGiuseppa} derived a new estimator for the covariance matrix under the complex Wishart model using inequalities relating the sought parameters.
In the following we derive new inequalities for this model.
Due to the major role of the covariance matrix in polarimetry~\cite{Conradsen2003}, we limit our analysis to inequalities that depend on $\boldsymbol{\Sigma}$.

Case~(ii) described in previous subsections paved the way for the new inequalities. 
The following results stem from the nonnegativity of the four distances:
\begin{equation} 
\operatorname{tr}(\boldsymbol{\Sigma}_2^{-1}\boldsymbol{\Sigma}_1+\boldsymbol{\Sigma}_1^{-1}\boldsymbol{\Sigma}_2)\geq 2p, 
\label{eq:b}
\end{equation}
\begin{align}
& \bigg(\frac{|\boldsymbol{\Sigma}_2|}{|\boldsymbol{\Sigma}_1|}\bigg)^{n\beta} |\boldsymbol{\Sigma}_2|^{-n} |(\beta \boldsymbol{\Sigma}_1^{-1}+(1-\beta)\boldsymbol{\Sigma}_2^{-1})^{-1}|^n  \nonumber\\
&+\bigg(\frac{|\boldsymbol{\Sigma}_1|}{|\boldsymbol{\Sigma}_2|}\bigg)^{n\beta} |\boldsymbol{\Sigma}_1|^{-n} |(\beta \boldsymbol{\Sigma}_2^{-1}+(1-\beta)\boldsymbol{\Sigma}_1^{-1})^{-1}|^n\leq 2, 
\label{eq:c}
\end{align}
\begin{equation}
 \frac{\log|\boldsymbol{\Sigma}_1|+\log|\boldsymbol{\Sigma}_2|}{2}\geq \log\bigg|\bigg(\frac{\boldsymbol{\Sigma}_1^{-1}+\boldsymbol{\Sigma}_2^{-1}}{2}\bigg)^{-1}\bigg|, \label{eq:d}
\end{equation}
and
\begin{equation}
\sqrt{|\boldsymbol{\Sigma}_1||\boldsymbol{\Sigma}_2|}\geq\bigg|\bigg(\frac{\boldsymbol{\Sigma}_1^{-1}+\boldsymbol{\Sigma}_2^{-1}}{2}\bigg)^{-1}\bigg|, 
\label{eq:e}
\end{equation}
respectively.
Fixing $\beta = 1/2$ in~\eqref{eq:c}, we obtain~\eqref{eq:e} directly; taking the logarithm of both sides of~\eqref{eq:e} yields~\eqref{eq:d}.
This result is justified by the following two relations: 
\begin{enumerate}
\item $d_{\text{B}}(\boldsymbol{\theta}_1,\boldsymbol{\theta}_2)=d_{\text{R}}^{1/2}(\boldsymbol{\theta}_1,\boldsymbol{\theta}_2)$ and
\item $d_{\text{H}}(\boldsymbol{\theta}_1,\boldsymbol{\theta}_2)=1-\exp{\bigl(-\frac12 {d_{\text{R}}^{1/2}(\boldsymbol{\theta}_1,\boldsymbol{\theta}_2)} \bigr)}$.
\end{enumerate}

The revised Wishart~\cite{ErsahinandCumming2010} and Bartlett distances~\cite{KerstenandLeeandAinsworth2005} can be obtained in a new and simple manner.
Indeed, the revised Wishart distance ($d_\text{RW}$) can be derived after simple manipulations of inequality~\eqref{eq:b}, yielding:
$$
d_\text{RW}(\boldsymbol{\Sigma}_1,\boldsymbol{\Sigma}_2)=\frac{\operatorname{tr}(\boldsymbol{\Sigma}_1\boldsymbol{\Sigma}_2^{-1}+\boldsymbol{\Sigma}_2\boldsymbol{\Sigma}_1^{-1})}{2}-p\geq 0.
$$
The Bartlett distance arises after taking the logarithm of both sides of inequality~\eqref{eq:e}.
Straightforward algebra leads to:
$$
\ln\frac{|\boldsymbol{\Sigma}_1+\boldsymbol{\Sigma}_2|^2}{|\boldsymbol{\Sigma}_1||\boldsymbol{\Sigma}_2|}-2\,p\,\ln 2\geq 0.
$$
The leftmost term in the inequality above is referred to as the Bartlett distance \cite{KerstenandLeeandAinsworth2005,ErsahinandCumming2010}.

\subsection{Finite sample size behavior}\label{sec:InfluenceEstimation}

We assessed the influence of estimation on the size of the new hypothesis tests using simulated data.
To that end, the study was conducted considering the following simulation parameters: number of looks $n=n_1=n_2\in\{4,8,16\}$ and the forest covariance matrix shown in~\eqref{matrixEX} with $x=360932$.
The sample sizes relate to square windows of size $7\times 7$, $11\times 11$, and $20\times 20$ pixels, i.e., $N_1,N_2 \in\{49,121,400\}$.
Nominal significance levels  $\alpha\in\{1\%,5\%\}$ were verified.

Let $T$ be the number of Monte Carlo replicas and $R$ the number of cases for which the null hypothesis is rejected at nominal level $\alpha$.
The empirical test size is given by $\widehat{\alpha}_{1-\alpha}={R}/{T}$ .
Following the methodology described in~\cite{HypothesisTestingSpeckledDataStochasticDistances}, we employed $T=5500$ replicas.

Table~\ref{table1} presents the empirical test sizes at $1\%$ and $5\%$ nominal levels, the execution time in milliseconds, the test statistic mean ($\overline{S}$), and coefficient of variation ($\text{CV}$).  
All numerical calculations and the execution time quantification were performed, running on a PC with an Intel Core 2 Duo processor \unit[2.10]{GHz}, \unit[4]{GB} of RAM, Windows XP, and the \texttt R platform v.~2.8.1.
For each case, the best obtained empirical sizes and distance means are in boldface.
Results for $N_1=49, N_2=\{121,400\}$ and $N_1=121, N_2=400$ are consistent with the ones shown, and are omitted for brevity.

We tested ten parameters: nine related to the covariance matrix of order $p=3$, and the number of looks $L$, leading to test statistics which asymptotically follow $\chi^2_{10}$ distributions. 
Thus, the statistics expected value should converge in probability to $10$, as a consequence of the weak law of large numbers.
In Table~\ref{table1}, notice that $\overline{S}$ tends to $10$ as the sample size increases. 
By fixing the sample size while varying the number of looks $n$, test sizes obey the inequalities $S_{\text{H}} \leq S_{\text{B}} \leq S_{\text{R}}^{\beta}\leq S_{\text{KL}}$, as illustrated in Fig.~\ref{CurvesWishart}.
These inequalities suggest that, for this study, the statistics based on the Kullback-Leibler distance is the best discrimination measure.

Regarding execution times, the Kullback-Leibler-based test presented the best performance,  while the test based on the Hellinger distance showed the best empirical test size in 6 out of 18 cases.

\begin{table*}[hbt]
\centering
\setlength{\tabcolsep}{2.5pt}                                                                                                        
\caption{Empirical sizes for B$_1$}\label{table1}
\begin{tabular}{c@{ }r@{ }r  r@{\quad }r@{\,\,\,}c@{}r@{\,\,\,\,\,}r  r@{\quad }r@{\,\,\,}c@{}r@{\,\,\,\,\,}r r@{\quad}r@{\,\,\,}c@{}r@{\,\,\,\,\,}r}\toprule
\multicolumn{3}{c}{Factors}& \multicolumn{5}{c}{$n=4$} & \multicolumn{5}{c}{$n=8$} & \multicolumn{5}{c}{$n=16$} \\
\cmidrule(lr{.25em}){4-8} \cmidrule(lr{.25em}){9-13} \cmidrule(lr{.25em}){14-18}
$S_{\phi}^h$ & $N_1$ & $N_2$ & $1\%$ & $5\%$ & $\text{time (ms)}$ & $\overline{S}$ &  $\text{CV}$
& $1\%$ & $5\%$ & $\text{time (ms)}$ & $\overline{S}$ &  $\text{CV}$
& $1\%$ & $5\%$ & $\text{time (ms)}$ & $\overline{S}$ &  $\text{CV}$ \\
\cmidrule(lr{.25em}){1-3} \cmidrule(lr{.25em}){4-8} \cmidrule(lr{.25em}){9-13} \cmidrule(lr{.25em}){14-18}
$S_{\text{KL}}$
&49& 49 &  1.309 & 5.491 &  0.44 & 10.189 & 44.804 &   1.472 & 6.291   & 0.49 &   10.292 & 45.873  &  1.509 & 6.364    & 0.39 &  10.400 & 45.828 \\
&121&121&  \textbf{0.818} & \textbf{4.545} &  0.43 &  9.843 & 44.743 &   1.255 & 5.618   & 0.49 &   10.052 & 45.241  &  \textbf{1.000} & \textbf{4.836}    & 0.48 &   9.982 & 44.309 \\
&400&400&  1.055 & \textbf{4.836} &  0.42 &  9.950 & 45.272 &   1.055 & 5.327   & 0.50 &   10.073 & 44.895  &  1.036 & \textbf{4.655}    & 0.46 &  10.051 & 44.539 \\
\cmidrule(lr{.25em}){1-3} \cmidrule(lr{.25em}){4-8} \cmidrule(lr{.25em}){9-13} \cmidrule(lr{.25em}){14-18}
$S_{\text{R}}^{\beta}$
&49& 49 &  1.255 & 5.309 &  0.54 & 10.157 & 44.658 &  1.436 & 6.127 & 0.51 & 10.272 & 45.782  &  1.473 &  6.291 & 0.58 & 10.385 & 45.763 \\
&121&121&  0.800 & 4.473 &  0.57 &  9.830 & 44.687 &  1.236 & 5.582 & 0.61 & 10.044 & 45.203  &  0.964 &  4.800 & 0.56 &  9.976 & 44.286 \\
&400&400&  \textbf{1.036} & 4.836 &  0.57 &  9.946 & 45.255 &  1.054 & 5.327 & 0.57 & 10.071 & 44.885  &  1.036 &  4.655 & 0.65 & 10.049 & 44.531 \\
\cmidrule(lr{.25em}){1-3} \cmidrule(lr{.25em}){4-8} \cmidrule(lr{.25em}){9-13} \cmidrule(lr{.25em}){14-18}
$S_{\text{B}}$%
&49& 49 &  \textbf{1.164} & \textbf{5.055} &  1.10 & 10.101 & 44.408 &  1.418  & 5.873 & 1.12 & 10.235 & 45.624  &  1.436 &  6.164 & 1.07 & 10.358 & 45.650 \\
&121&121&  0.782 & 4.418 &  1.11 &  9.809 & 44.588 &  1.218  & 5.473 & 0.99 & 10.030 & 45.135  &  0.963 &  4.745 & 1.12 &  9.967 & 44.245 \\
&400&400&  \textbf{1.036} & 4.836 &  0.99 &  9.939 & 45.224 &  1.055  & 5.323 & 1.13 & 10.066 & 44.866  &  1.036 &  4.636 & 1.20 & 10.046 & 44.515 \\
\cmidrule(lr{.25em}){1-3} \cmidrule(lr{.25em}){4-8} \cmidrule(lr{.25em}){9-13} \cmidrule(lr{.25em}){14-18}
$S_{\text{H}}$
&49& 49 &  0.655 & 3.891 &  1.10 &  9.797 & 43.017  & \textbf{0.891} & \textbf{4.400} & 1.12 &   9.920 & 44.184  &  \textbf{1.000}  &  \textbf{4.782} & 1.07   & 10.035 & 44.225 \\
&121&121&  0.618 & 4.018 &  1.11 &  9.691 & 44.054  &  \textbf{0.927} & \textbf{5.018} & 0.99 &   9.906 & 44.580  &  0.855  &  4.091 & 1.12   &  9.845 & 43.705 \\
&400&400&  \textbf{1.036} & 4.691 &  0.99 &  9.902 & 45.056  &  \textbf{0.909} & \textbf{5.145} & 1.13 &  10.029 & 44.698  &  \textbf{1.018}  &  4.473 & 1.20   & 10.009 & 44.348 \\%
\bottomrule
\end{tabular}
\end{table*}

The presented methodology for assessing test sizes  was also applied to the three forest samples from the E-SAR image shown in Fig.~\ref{fig0}.
Each sample was submitted to the following procedure~\cite{HypothesisTestingSpeckledDataStochasticDistances}: 
\begin{itemize}
\item[(i)] split the sample in disjoint blocks of size $N_1$;
\item[(ii)] for each block from~(i), split the remaining sample in disjoint blocks of size $N_2$;
\item[(iii)] perform the hypothesis test as described in Proposition~\ref{p-3} for each pair of samples with sizes $N_1$ and $N_2$.
\end{itemize}

Table~\ref{table2} presents the results, omitting some entries as in Table~\ref{table1}. 
All test sizes were smaller than the nominal level, i.e., the proposed tests do not reject the null hypothesis when similar samples are considered.           

\begin{figure}[htb]
\centering
{\includegraphics[width=.8\linewidth]{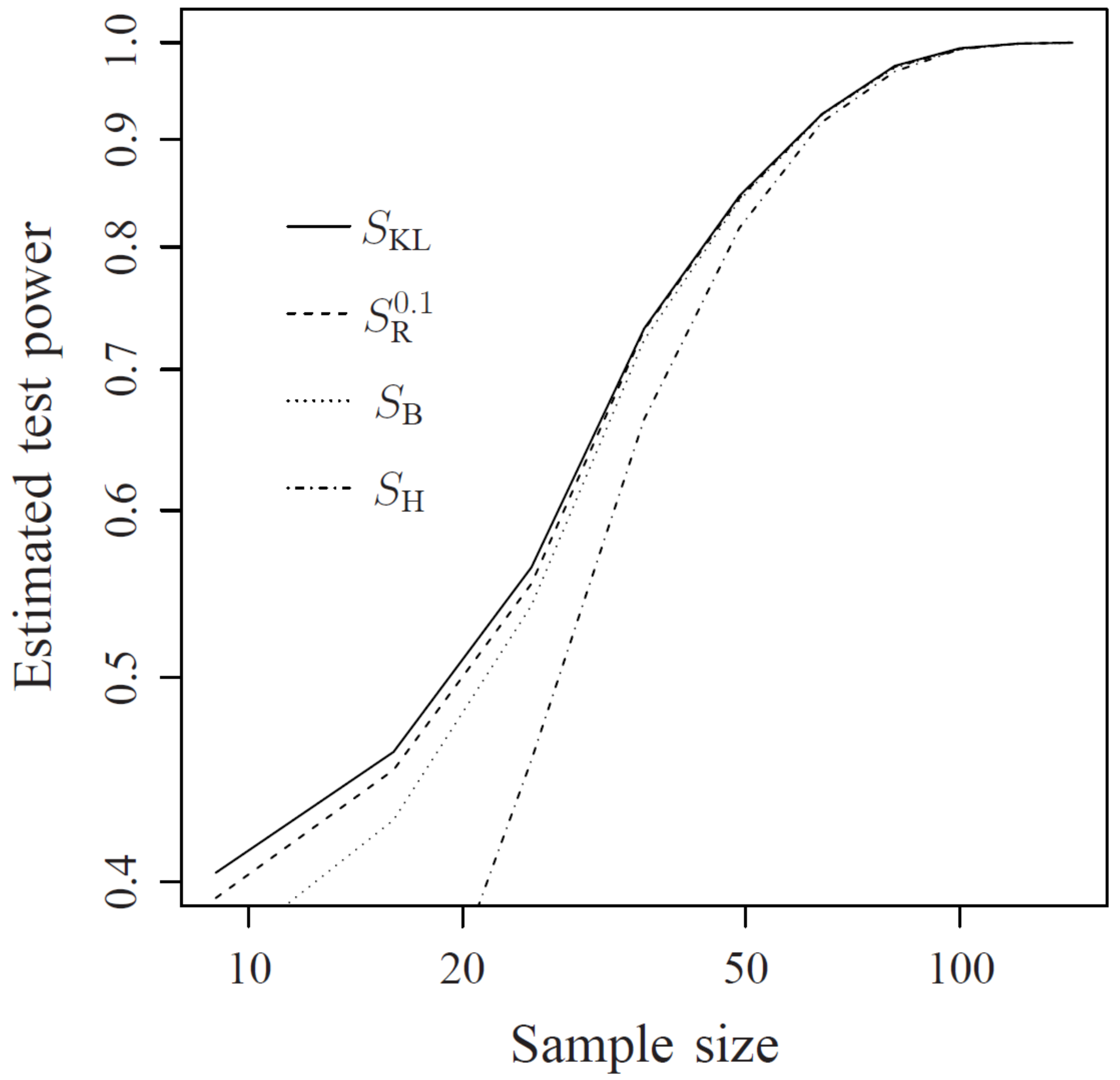}}
\caption{Empirical test power in semilogarithmic scale.}
\label{otherpower}
\end{figure}

We also made the following study on the tests power: In each one of $T$ Monte Carlo experiments, random matrices both of sizes $N\in\{9,16,25,36,49,64,81,100,121,144\}$ were sampled from the $\mathcal{W_R}(\boldsymbol{\Sigma}(360932),4)$ and from the $\mathcal{W_R}(\boldsymbol{\Sigma}(360932)\cdot (1+0.2),4)$ distributions.
The covariance matrix $\boldsymbol{\Sigma}(x)$ is given in~\eqref{matrixEX}, and the experiment consists in contrasting samples from the relaxed Wishart distribution it indexes, and the law indexed by a version scaled by $1.2$, arbitrarily chosen. 
Subsequently, it was verified whether these samples come from similar populations according to Proposition~\ref{p-3}.
Let $R$ be the number of situations for which the null hypothesis is rejected at nominal level $\alpha$; the empirical test power is given by $R/T$.
Fig.~\ref{otherpower} presents these estimates for the test power. 
Notice that the discrimination ability is about the same for all tests above $N=49$.

In general terms, the proposed hypothesis tests presented good results regarding their power even for small samples: with samples of size $49$, they are able to discriminate between covariance matrices which are only $20\%$ different in about $80\%$ of the time.
As the sample size increases, all the tests discriminate better and better.                                                                                        

\begin{table*}[hbt]
\centering
\caption{Empirical sizes for forests}\label{table2}
\begin{tabular}{*{3}{p{6pt}}rrr@{}l@{}r  rrr@{}l@{}r  rrr@{}l@{}r}
\toprule
\multicolumn{3}{c}{Factors}& \multicolumn{5}{c}{B$_1$} & \multicolumn{5}{c}{B$_2$} & \multicolumn{5}{c}{B$_3$} \\
\cmidrule(lr{.25em}){1-3} \cmidrule(lr{.25em}){4-8} \cmidrule(lr{.25em}){9-13} \cmidrule(lr{.25em}){14-18}
$S_{\phi}^h$&$N_1$&$N_2$ & $1\%$ & $5\%$ & \multicolumn{1}{c}{$\overline{S}$} & $(\times 10^{-1})$ & \multicolumn{1}{c}{CV}
& $1\%$ & $5\%$ & \multicolumn{1}{c}{$\overline{S}$}& $(\times 10^{-1})$ & \multicolumn{1}{c}{CV}                          
& $1\%$ & $5\%$ & \multicolumn{1}{c}{$\overline{S}$}& $(\times 10^{-1})$ & \multicolumn{1}{c}{CV}\\                        
 \cmidrule(lr{.25em}){1-3} \cmidrule(lr{.25em}){4-8} \cmidrule(lr{.25em}){9-13} \cmidrule(lr{.25em}){14-18}                
$S_{\text{KL}}$
&49& 49 & 0.00 & 0.00 & \multicolumn{2}{c}{59.80}     & 157.36 &   0.00 & 0.00 & \multicolumn{2}{c}{45.16}  & 64.18 &  0.00 & 0.00 & \multicolumn{2}{c}{47.05} & 87.32 \\
&121&121& 0.00 & 0.00 & \multicolumn{2}{c}{38.27}     & 100.85 &   0.00 & 0.00 & \multicolumn{2}{c}{19.76}  & 61.66 &  0.00 & 0.00 & \multicolumn{2}{c}{29.54} & 87.72 \\
\cmidrule(lr{.25em}){1-3} \cmidrule(lr{.25em}){4-8} \cmidrule(lr{.25em}){9-13} \cmidrule(lr{.25em}){14-18}
$S_{\text{R}}^{\beta}$
&49& 49 & 0.00 & 0.00 & \multicolumn{2}{c}{40.77}  & 153.99 &   0.00 & 0.00 & \multicolumn{2}{c}{31.40} & 63.83 &  0.00 & 0.00 & \multicolumn{2}{c}{32.61} & 86.42 \\  
&121&121& 0.00 & 0.00 & \multicolumn{2}{c}{26.56}  & 100.41 &   0.00 & 0.00 & \multicolumn{2}{c}{13.79} & 61.51 &  0.00 & 0.00 & \multicolumn{2}{c}{20.54} & 87.17 \\  
\cmidrule(lr{.25em}){1-3} \cmidrule(lr{.25em}){4-8} \cmidrule(lr{.25em}){9-13} \cmidrule(lr{.25em}){14-18}
$S_{\text{B}}$
&49& 49 & 0.00 & 0.00 & \multicolumn{2}{c}{41.92} & 148.92 &   0.00 & 0.00 & \multicolumn{2}{c}{33.25} & 63.22 &  0.00 & 0.00 & \multicolumn{2}{c}{34.34} & 84.96 \\   
&121&121& 0.00 & 0.00 & \multicolumn{2}{c}{28.05} &  99.66 &   0.00 & 0.00 & \multicolumn{2}{c}{14.70} & 61.23 &  0.00 & 0.00 & \multicolumn{2}{c}{21.77} & 86.24 \\   
\cmidrule(lr{.25em}){1-3} \cmidrule(lr{.25em}){4-8} \cmidrule(lr{.25em}){9-13} \cmidrule(lr{.25em}){14-18}
$S_{\text{H}}$
&49& 49 & 0.00 & 0.00 & \multicolumn{2}{c}{36.51} & 132.59 &   0.00 & 0.00 & \multicolumn{2}{c}{31.61} & 60.27 &  0.00 & 0.00 & \multicolumn{2}{c}{32.22} & 79.05 \\   
&121&121& 0.00 & 0.00 & \multicolumn{2}{c}{26.41} &  96.60 &   0.00 & 0.00 & \multicolumn{2}{c}{14.38} & 59.97 &  0.00 & 0.00 & \multicolumn{2}{c}{20.89} & 82.53 \\   
\bottomrule
\end{tabular}
\end{table*}

\section{Applications}\label{sec:Applications}

This section presents two applications of the tests based on stochastic distances.
Firstly, a discrimination analysis was performed in order to assess the influence of image texture on the tests.
It is known that the complex Wishart distribution is more appropriated for describing homogeneous regions.
However, other polarimetric distributions, potentially more apt to describing textured areas, yield intractable expressions which depend on special functions, such as the hypergeometric and modified Bessel functions.
In order to quantify textures, we considered distances between relaxed scaled complex Wishart laws as proposed by Anfinsen~\textit{et~al.}~\cite{AnfinsenJenssenEltoft2009}. 
Secondly, stochastic distances are embedded into the $k$-means method in order to identify groups in PolSAR data. 
The performance of four distances was assessed by means of a synthetic image generated from the relaxed scaled complex Wishart distribution.

\subsection{Discrimination analysis}\label{sec:Applications1}

\begin{figure}[htb]
\centering
\includegraphics[width=.7\linewidth]{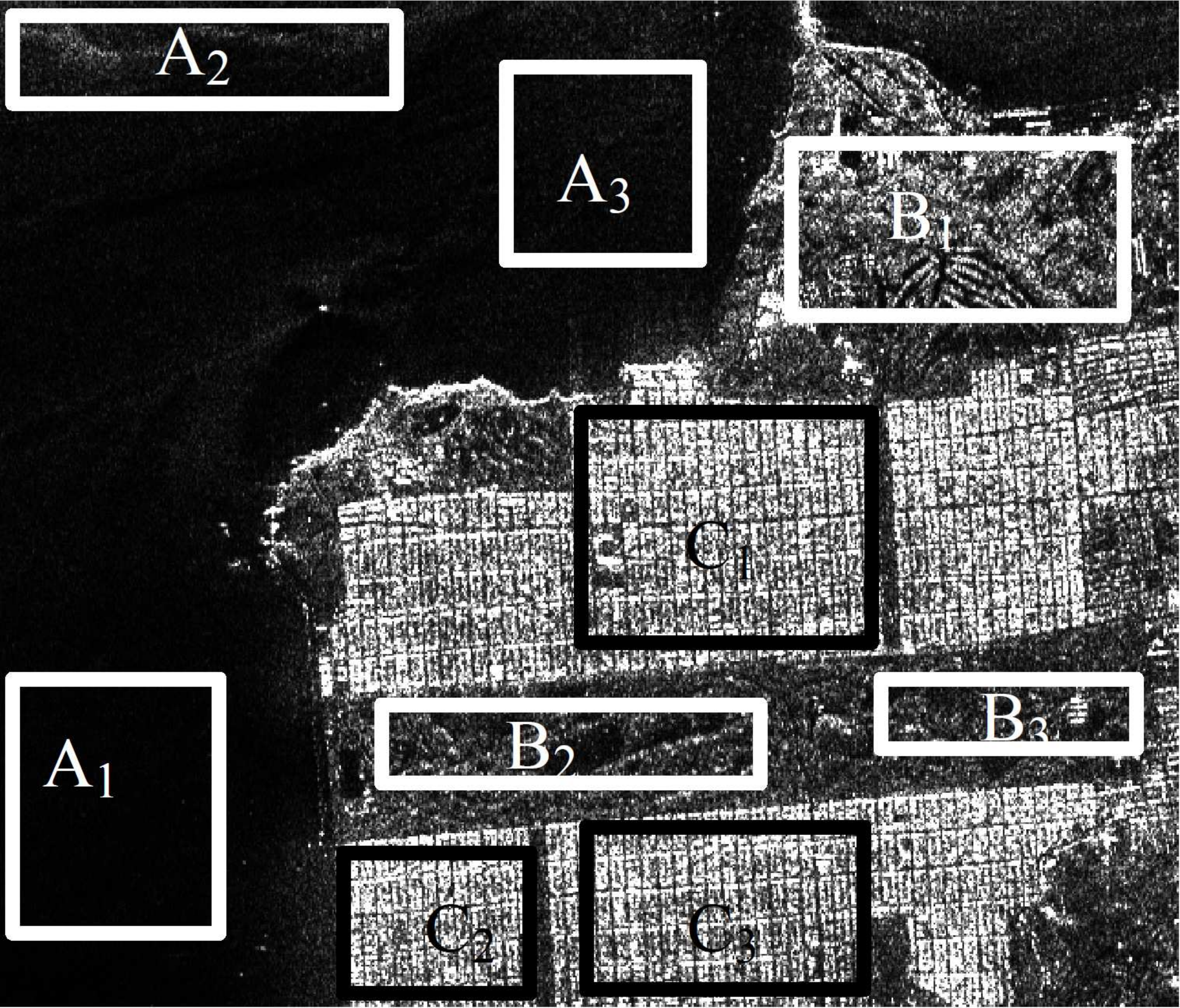}
\caption{AIRSAR HH data and samples}
\label{SanFrancisco1}
\end{figure}

AIRSAR was an airborne mission with PolSAR capabilities, designed and built by the Jet Propulsion Laboratory, which operated at P-, L-, and C-bands~\cite{ESA2009}.
Fig.~\ref{SanFrancisco1} shows a $550 \times 645$ pixels image (HH channel) of San Francisco recorded by this sensor, acquired with four nominal looks.
Nine areas were chosen to represent three different degrees of roughness: homogeneous, heterogeneous, and extremely heterogeneous, labeled as $\text{A}_{i}$, $\text{B}_{i}$, and $\text{C}_{i}$, respectively, for ${i}=1,2,3,$.

The parameters of the complex Wishart distribution were estimated by maximum likelihood, cf.\ Eq.~\eqref{escorefunction}.
Table~\ref{tableesti} presents the estimated number of looks and determinants of the complex covariance matrices for each area, along with the number of observations.

Goodman~\cite{Goodmana} studied the distribution of the determinant of the complex covariance matrix, which is clasically understood as a generalized variance.
In PolSAR, this quantity is related to the speckle variability, defined as the effect of the speckle noise resulting from multipath interference.
Additionally, when there is variability due to texture it is caused by the spatial variability of the reflectance, and it is understood as ``heterogeneity'' or ``roughness''.
This source of variability can be captured by, for instance, the roughness parameter of the polarimetric $\mathcal G^0$ law~\cite{FreitasFreryCorreia:Environmetrics:03,FreryCorreiaFreitas:ClassifMultifrequency:IEEE:2007}.

We observed that the elements of the covariance matrices become larger along with the determinant when the heterogeneity increases.
The most homogeneous region, $\text{A}_{1}$, has the covariance matrix with smallest determinant.
Sample $\text{B}_1$ has the largest determinant among heterogeneous regions.
This suggests that this last sample is the most heterogeneous, even with the presence of a double bounce~\cite{ParametricNonparametricTestsSpeckledImagery} in sample $\text{B}_3$.
Urban areas (labeled $\text{C}_{1}$, $\text{C}_{2}$ and $\text{C}_{3}$ in Fig.~\ref{SanFrancisco1}), which are extremely heterogeneous targets, lead to the largest determinants.
Additionally, the estimated number of looks decreases with the heterogeneity.


\begin{table}[htb]                                                                                                                 
\centering                 
\setlength{\tabcolsep}{2.5pt}                                                                                                        
\caption{Estimated number of looks and generalized variance}\label{tableesti}
\begin{tabular}{c@{ }ccccc} \toprule
\multicolumn{2}{c}{} & \multicolumn{3}{c}{Subscript of regions} \\ \cmidrule(lr{.25em}){3-5}
Estimates & {Region}  & \multicolumn{1}{c}{${i}=1$}&\multicolumn{1}{c}{${i}=2$}&\multicolumn{1}{c}{${i}=3$}\\ 
\cmidrule(lr{.25em}){3-5}
$\widehat{n} $ & $\text{A}_{i}$  &                                                    
4.04 & 3.75 & 3.98 \\                                                                 
$|\widehat{\boldsymbol{\Sigma}}|$& &                                                  
$3.24\times {10}^{{-9}}$ & $59.71\times {10}^{{-9}}$ & $18.83\times {10}^{{-9}}$\\    
\# pixels & &                                                                         
15960 & 10339 &  11449 \\ \cmidrule(lr{.25em}){3-5}                                   
& $\text{B}_{i}$  &                                                                   
3.05 & 3.21  & 3.15  \\                                                               
&& $35.86\times {10}^{{-5}}$ & $7.38\times {10}^{{-5}}$ & $10.87\times {10}^{{-5}}$ \\
&  & 17385 & 9152 & 5499 \\ \cmidrule(lr{.25em}){3-5}                                 
& $\text{C}_{i}$ & 3.03 & 3.12  & 3.09 \\                                             
&&  $1.97\times {10}^{{-3}}$ & $1.46\times {10}^{{-3}}$ & $1.17\times {10}^{{-3}}$ \\ 
& & 20320 & 8034 &  13770 \\                                                          
\bottomrule                                              
\end{tabular}                                                                                                                      
\end{table}

Stochastic distances were computed between pairs of these estimated distributions.
Table~\ref{tableestimationMLA} shows the distances between regions of the same class.
In all but one case, the values were found to be ordered as follows: $d_\text{KL}>d_\text{R}^{\text{0.9}}\geq d_\text{B} \geq d_\text{H} > d_\text{R}^{\text{0.1}}$.
The only discrepancy occurs when comparing homogeneous regions $\text{A}_1$ and $\text{A}_2$, where the last inequality is not preserved.

\begin{table}[htb]                                               
\centering                                                       
\caption{Distances between regions of similar roughness}         
\label{tableestimationMLA}                                       
\begin{tabular}{crrrrr}\toprule                                  
Regions&  $d_\text{KL}$ & $d_\text{R}^{0.9}$ &                   
$d_\text{B}$ & $d_\text{H}$ & $d_\text{R}^{0.1}$ \\ \midrule     
A$_1$-A$_2$ & 19.83 & 13.13 & 2.66 & 0.93 & 1.46   \\            
A$_1$-A$_3$ &  7.34 &  5.91 & 1.41 & 0.76 & 0.66   \\            
A$_2$-A$_3$ &  2.01 &  1.73 & 0.45 & 0.36 & 0.19   \\ \midrule   
B$_1$-B$_2$ &  1.83 &  1.58 & 0.41 & 0.34 & 0.18   \\            
B$_1$-B$_3$ &  1.11 &  0.97 & 0.26 & 0.23 & 0.11   \\            
B$_2$-B$_3$ &  0.26 &  0.23 & 0.06 & 0.06 & 0.03   \\ \midrule   
C$_1$-C$_2$ &  0.35 &  0.31 & 0.09 & 0.08 & 0.03   \\            
C$_1$-C$_3$ &  0.21 &  0.18 & 0.05 & 0.05 & 0.02   \\            
C$_2$-C$_3$ &  0.19 &  0.17 & 0.05 & 0.05 & 0.02   \\ \bottomrule
\end{tabular}                                                    
\end{table}

Table~\ref{tableestimationMLB} presents the distances between regions of different roughness.
Similarly to the univariate case~\cite{HypothesisTestingSpeckledDataStochasticDistances}, in these cases regions become more distinguishable.
In all cases, the distances satisfy
$$
d_{\mathcal M}\bigl[(\widehat{\Sigma}_k,\widehat n),(\widehat{\Sigma}_\ell,\widehat n)\bigr]<d_{\mathcal M}\bigl[(\widehat{\Sigma}_k,\widehat n),(\widehat{\Sigma}_m,\widehat n)\bigr],
$$
if $||\widehat{\Sigma}_k|-|\widehat{\Sigma}_\ell||<||\widehat{\Sigma}_k|-|\widehat{\Sigma}_m||$.

\begin{table}[hbt]                                                                    
\centering                                                                            
\caption{Distances between regions of different roughness}\label{tableestimationMLB}  
\begin{tabular}{crrrrr}\toprule                                                       
Regions & $d_\text{KL}$ & $d_\text{R}^{0.9}$ &                                        
$d_\text{B}$ & $d_\text{H}$ & $d_\text{R}^{0.1}$ \\ \midrule                          
A$_1$-B$_1$ &  621.16 &  91.09 & 12.38 & 1.00 & 10.12	 \\                             
A$_1$-B$_2$ &  318.09 &  74.13 & 10.61 & 1.00 &  8.24	 \\                             
A$_1$-B$_3$ &  397.15 &  79.29 & 11.08 & 1.00 &  8.81	 \\                             
A$_2$-B$_1$ &  222.93 &  60.04 &  8.72 & 1.00 &  6.67	 \\                             
A$_2$-B$_2$ &  119.27 &  46.15 &  7.23 & 1.00 &  5.13	 \\                             
A$_2$-B$_3$ &  152.90 &  51.65 &  7.81 & 1.00 &  5.74	 \\                             
A$_3$-B$_1$ &  361.81 &  72.51 & 10.14 & 1.00 &  8.06	 \\                             
A$_3$-B$_2$ &  185.07 &  57.19 &  8.50 & 1.00 &  6.35	 \\                             
A$_3$-B$_3$ &  239.22 &  62.80 &  9.05 & 1.00 &  6.98	 \\ \midrule                    
A$_1$-C$_1$ & 1559.32 & 110.64 & 14.65 & 1.00 & 12.29	 \\                             
A$_1$-C$_2$ & 1469.62 & 109.18 & 14.57 & 1.00 & 12.13	 \\                             
A$_1$-C$_3$ & 1393.29 & 106.23 & 14.16 & 1.00 & 11.80	 \\                             
A$_2$-C$_1$ &  621.61 &  77.63 & 10.70 & 1.00 &  8.62	 \\                             
A$_2$-C$_2$ &  575.70 &  75.85 & 10.56 & 1.00 &  8.43	 \\                             
A$_2$-C$_3$ &  554.52 &  74.31 & 10.29 & 1.00 &  8.26	 \\                             
A$_3$-C$_1$ & 1010.30 &  90.89 & 12.26 & 1.00 & 10.10	 \\                             
A$_3$-C$_2$ &  954.88 &  89.29 & 12.14 & 1.00 &  9.92	 \\                             
A$_3$-C$_3$ &  907.62 &  87.21 & 11.81 & 1.00 &  9.69	 \\ \midrule                    
B$_1$-C$_1$ &    4.76 &   3.91 &  0.95 & 0.61 &  0.43	  \\                            
B$_1$-C$_2$ &    4.23 &   3.54 &  0.88 & 0.59 &  0.39	  \\                            
B$_1$-C$_3$ &    3.88 &   3.25 &  0.81 & 0.55 &  0.36	  \\                            
B$_2$-C$_1$ &   13.41 &   9.50 &  2.02 & 0.87 &  1.06	  \\                            
B$_2$-C$_2$ &   12.44 &   8.97 &  1.93 & 0.86 &  1.00	  \\                            
B$_2$-C$_3$ &   11.18 &   8.10 &  1.75 & 0.83 &  0.90	  \\                            
B$_3$-C$_1$ &    9.42 &   7.24 &  1.65 & 0.81 &  0.80	  \\                            
B$_3$-C$_2$ &    8.71 &   6.79 &  1.57 & 0.79 &  0.75	  \\                            
B$_3$-C$_3$ &    7.59 &   5.96 &  1.38 & 0.75 &  0.66	  \\ \bottomrule                
\end{tabular}                                                                         
\end{table}

As expected, distances between samples of different classes are much larger than those between samples with similar roughness.

\subsection{Clustering with stochastic distances}

A common characteristic of segmentation and classification algorithms is their sensitivity to the dissimilarity measure they employ~\cite{ErsahinandCumming2010,Doulgerisetal2011}. 
As already presented, stochastic distances present good discriminatory properties and, therefore, can be used for identifying clusters in PolSAR data.
To that end, in the following we use the $k$-means method with these measures applied to, firstly, synthetic data, and, secondly, to a PolSAR image.

Consider $N$ observed covariance matrices $\boldsymbol{Z}_i$, $1\leq i\leq N$, and assume that each observation belongs to a class $\mathcal H_\ell$, $1\leq\ell\leq k$, with $k$ known.
Each class can be characterized by an unknown centroid $\boldsymbol{C}_\ell$, $1\leq \ell\leq k$, and the task is assigning each observation to a single class.
Algorithm~\ref{alg:kmeans} performs this task using stochastic distances as dissimilarity criteria.

\begin{algorithm}[hbt]
\caption{$k$-means using stochastic distances}\label{alg:kmeans}
\begin{algorithmic}[1]
\State Choose a set of arbitrary initial $k$ centroids $\boldsymbol{\mathcal{C}}=\{\boldsymbol{C}_1,\boldsymbol{C}_2,\ldots,\boldsymbol{C}_k\}$.
\State\label{algo1} For each $i\in\{1,2,\ldots,k\}$, set the cluster $ \boldsymbol{\mathcal F}_i$ as a set of pixels which are closer to $\boldsymbol{C}_i$ than to $\boldsymbol{C}_j$ (for all $i\neq j$) according to the rule
\begin{align*}
\boldsymbol{\mathcal F}_i=\{\boldsymbol{Z}_i\colon &d_{\mathcal M}([\boldsymbol{Z},n],[\boldsymbol{C}_i,n])\,\leq\\
& \min_{1\leq j \leq k,j\neq i } d_{\mathcal M}([\boldsymbol{Z},n],[\boldsymbol{C}_j,n])\},
\end{align*}
where $d_{\mathcal M}$ is a stochastic distance.
\State\label{algo2} Reset ${\boldsymbol{C}}_i$ as the sample mean of the elements of $\boldsymbol{\mathcal F}_i$ defined in the step~\ref{algo1}, for $i\in\{1,2,\ldots,k\}$.
\State Repeat steps \ref{algo1} and~\ref{algo2} until {$\boldsymbol{\mathcal{C}}$ no longer changes}.
In other words, when the following measure assumes zero value:
$$
{\mathcal H}(v)=\sum_{j=1}^{\text{\# pixels}}\,\sum_{i=1}^k\,\mid\mathbb{I}_{\boldsymbol{\mathcal F}_i}(x_{j,v})-\mathbb{I}_{\boldsymbol{\mathcal F}_i}(x_{j,v-1})\mid,
$$
for all $n\geq 1$, where $x_{j,v}$ represents the $j$th element of the vector labels associated to elements of the vectorization from data matrix at $v$th iteration, $x_{j,0}$ is $j$th label of Initial solution, and $\mathbb{I}_{\boldsymbol{\mathcal F}_i}(\cdot)$ is the indicator function of set $\boldsymbol{\mathcal F}_i$.
\end{algorithmic}
\end{algorithm}

\begin{figure}[htb]
\centering
\subfigure[Synthetic image~\label{cluster1}]{\includegraphics[width=.37\linewidth]{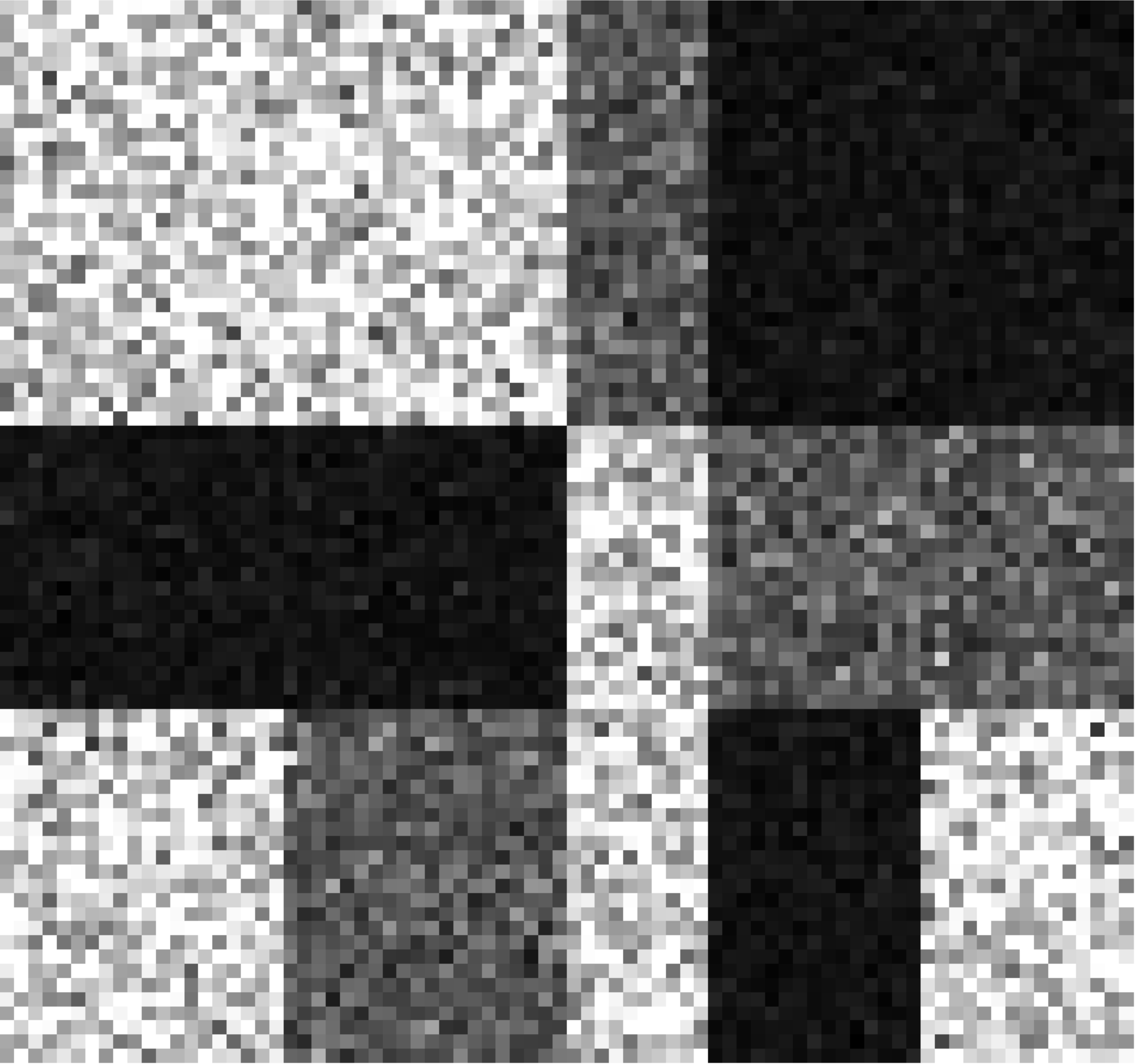}}
\subfigure[Initial solution~\label{cluster2}]{\includegraphics[width=.41\linewidth]{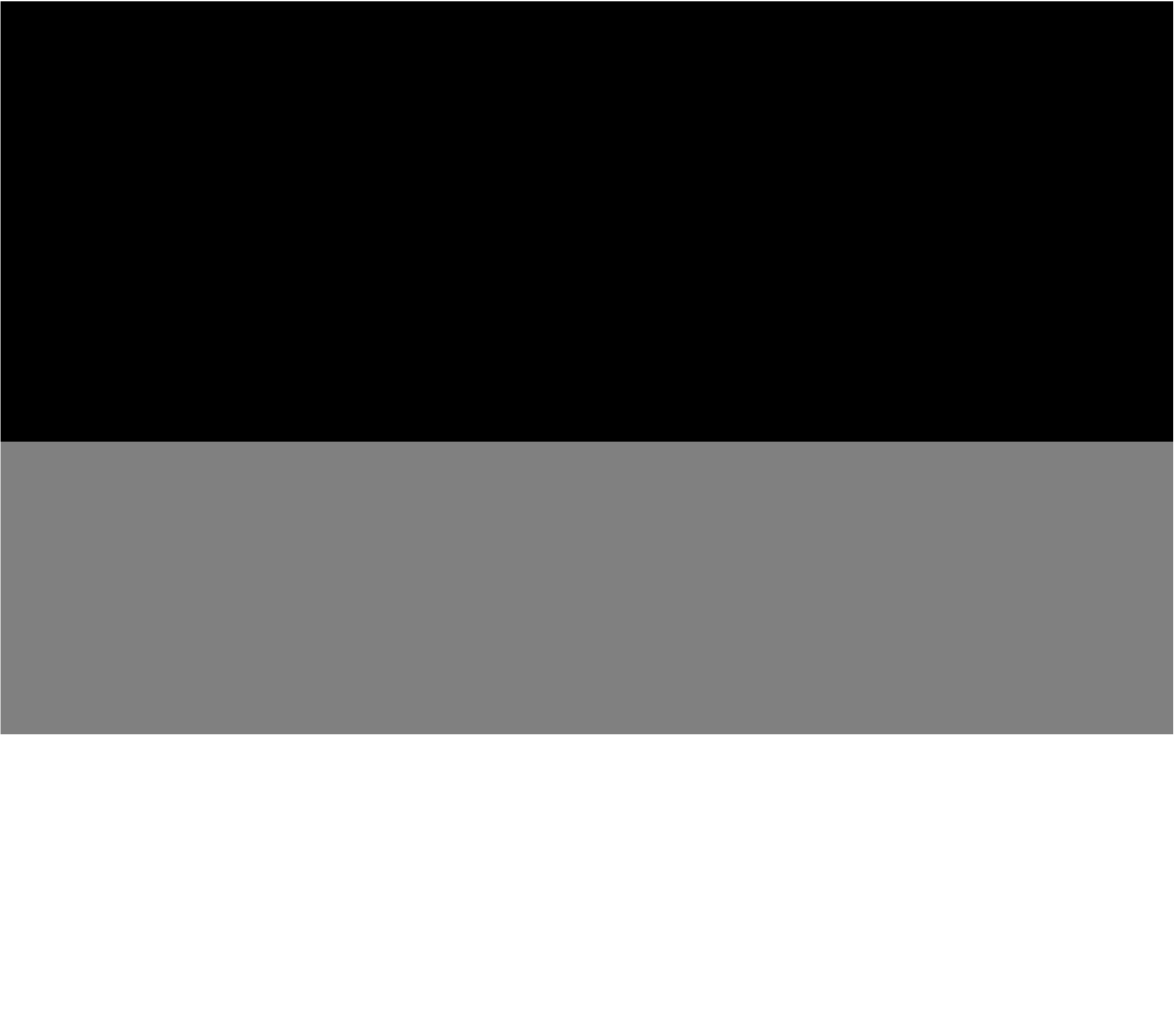}}
\subfigure[$d_\text{KL}$ clusters~\label{cluster3}]{\includegraphics[width=.38\linewidth]{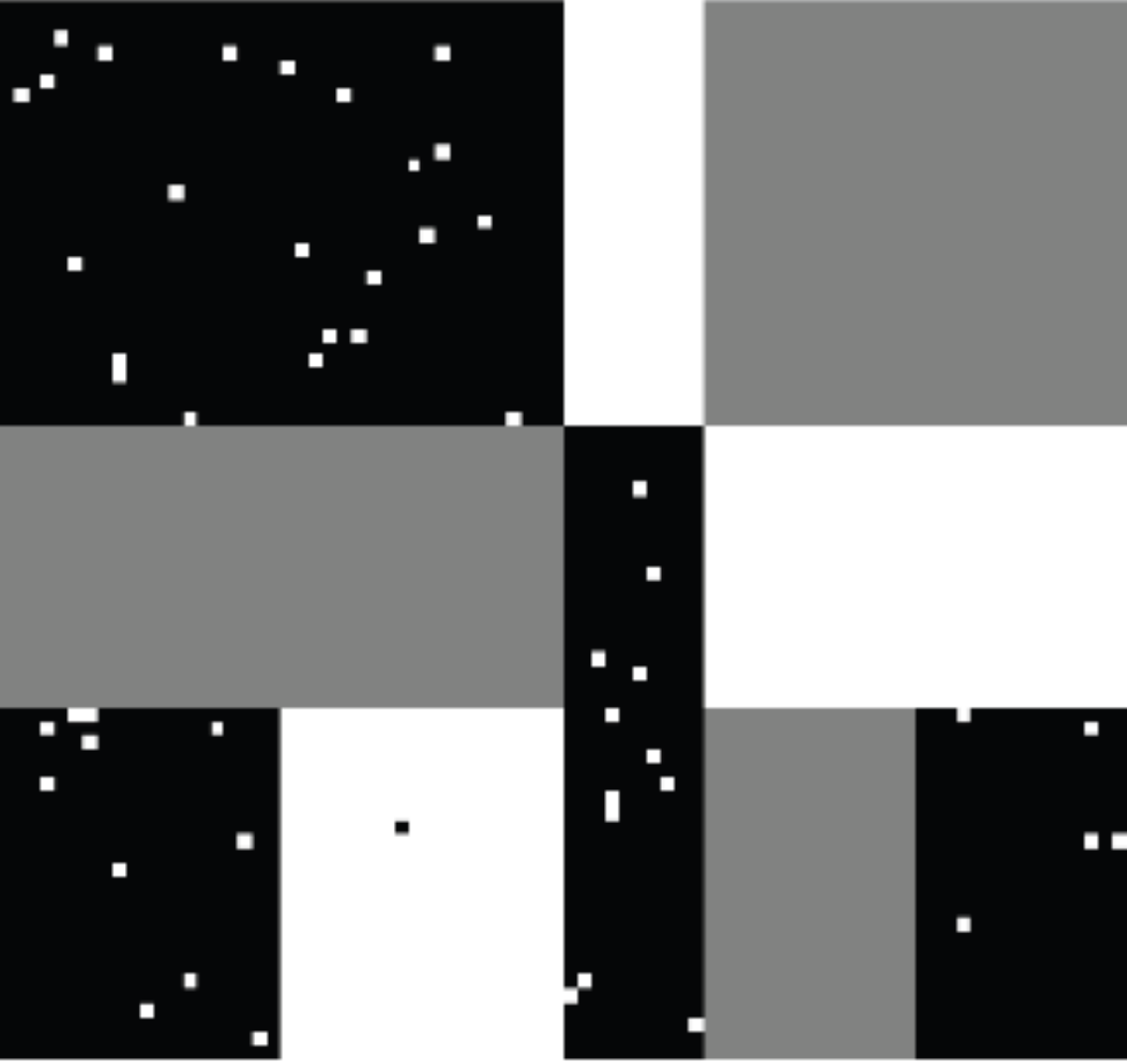}}
\subfigure[$d_\text{B}$  clusters~\label{cluster4}]{\includegraphics[width=.38\linewidth]{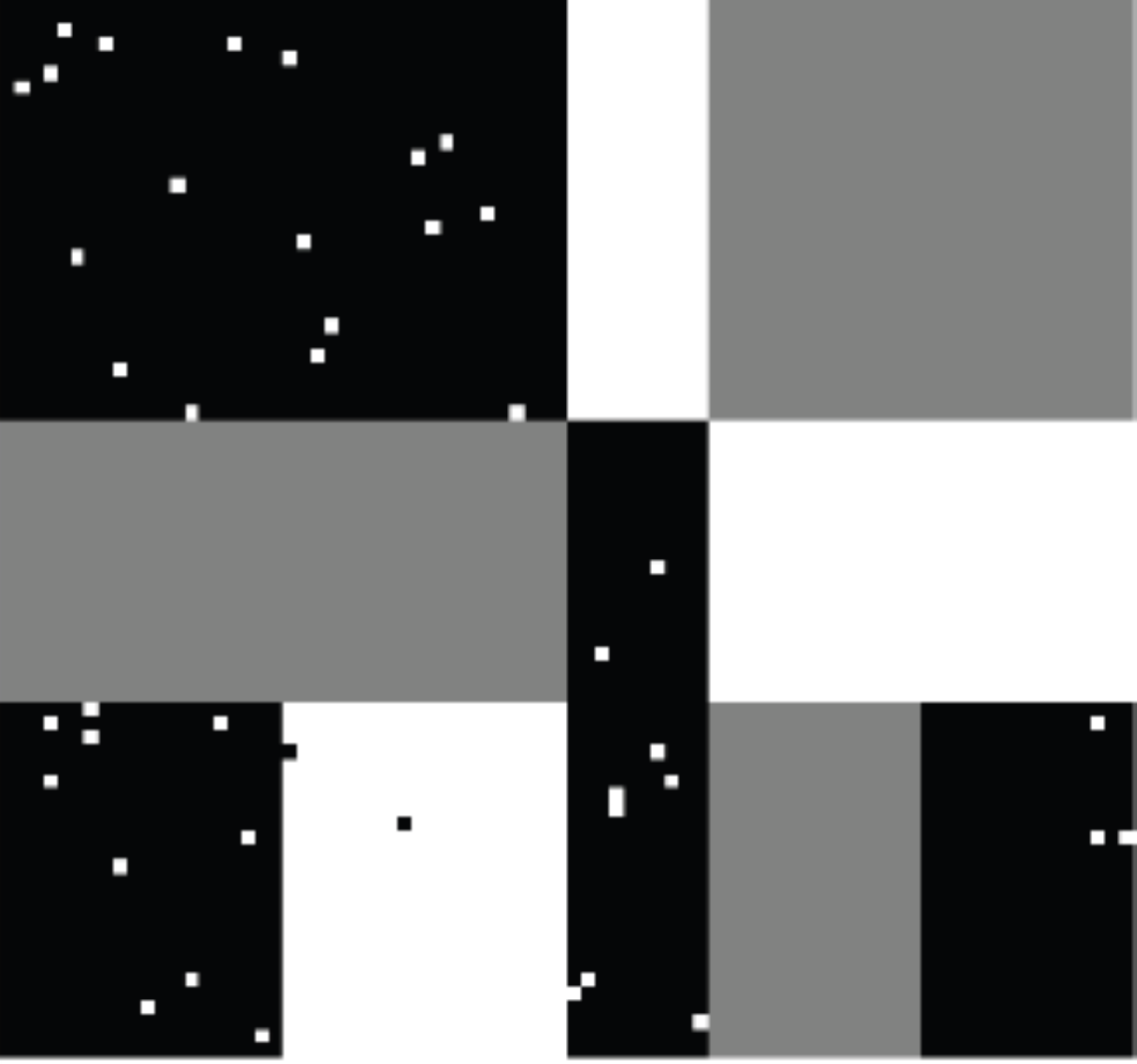}}
\subfigure[ $d_\text{H}$ clusters~\label{cluster5}]{\includegraphics[width=.38\linewidth]{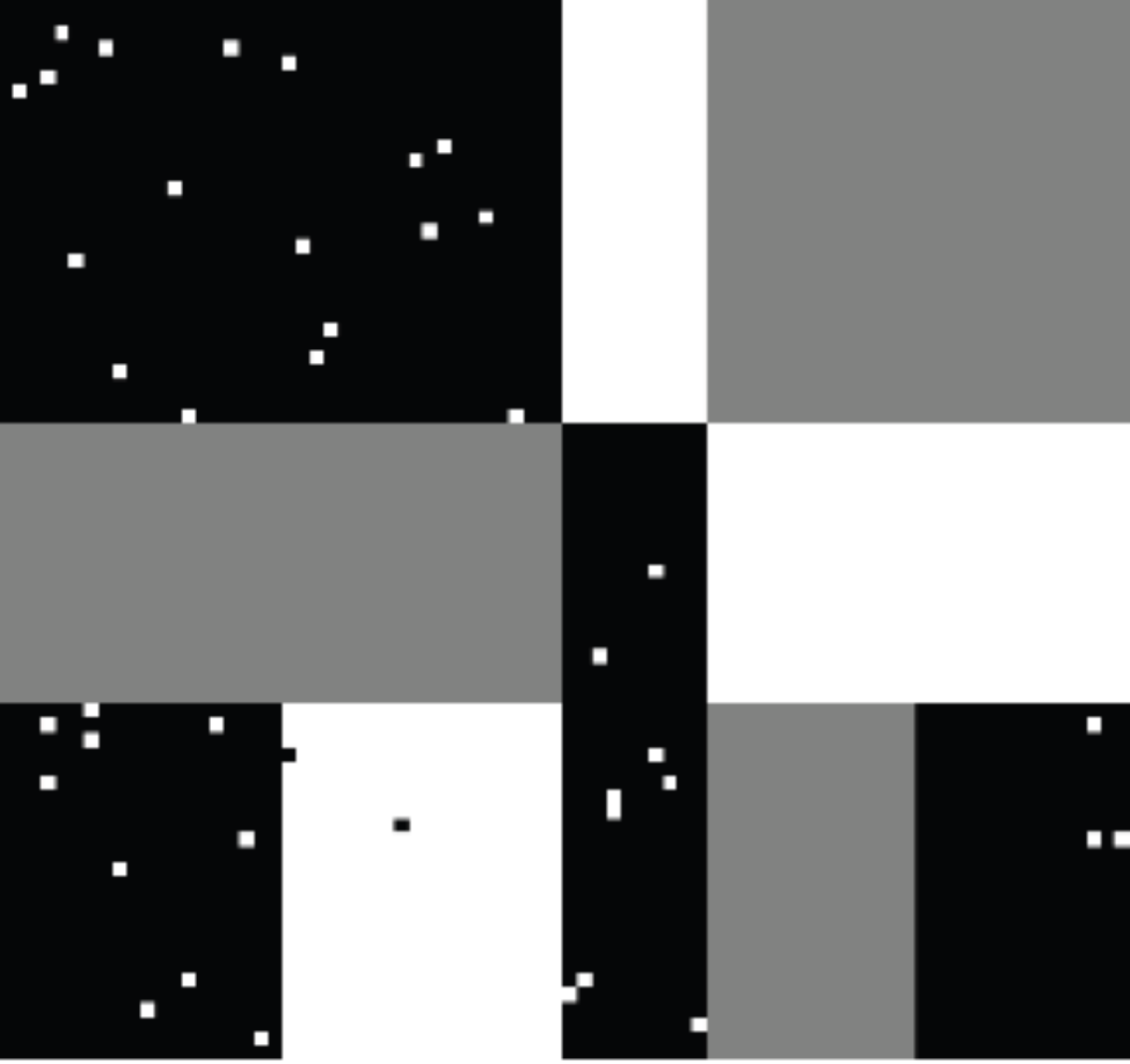}}
\subfigure[ $d_\text{R}^{0.1}$ clusters ~\label{cluster6}]{\includegraphics[width=.38\linewidth]{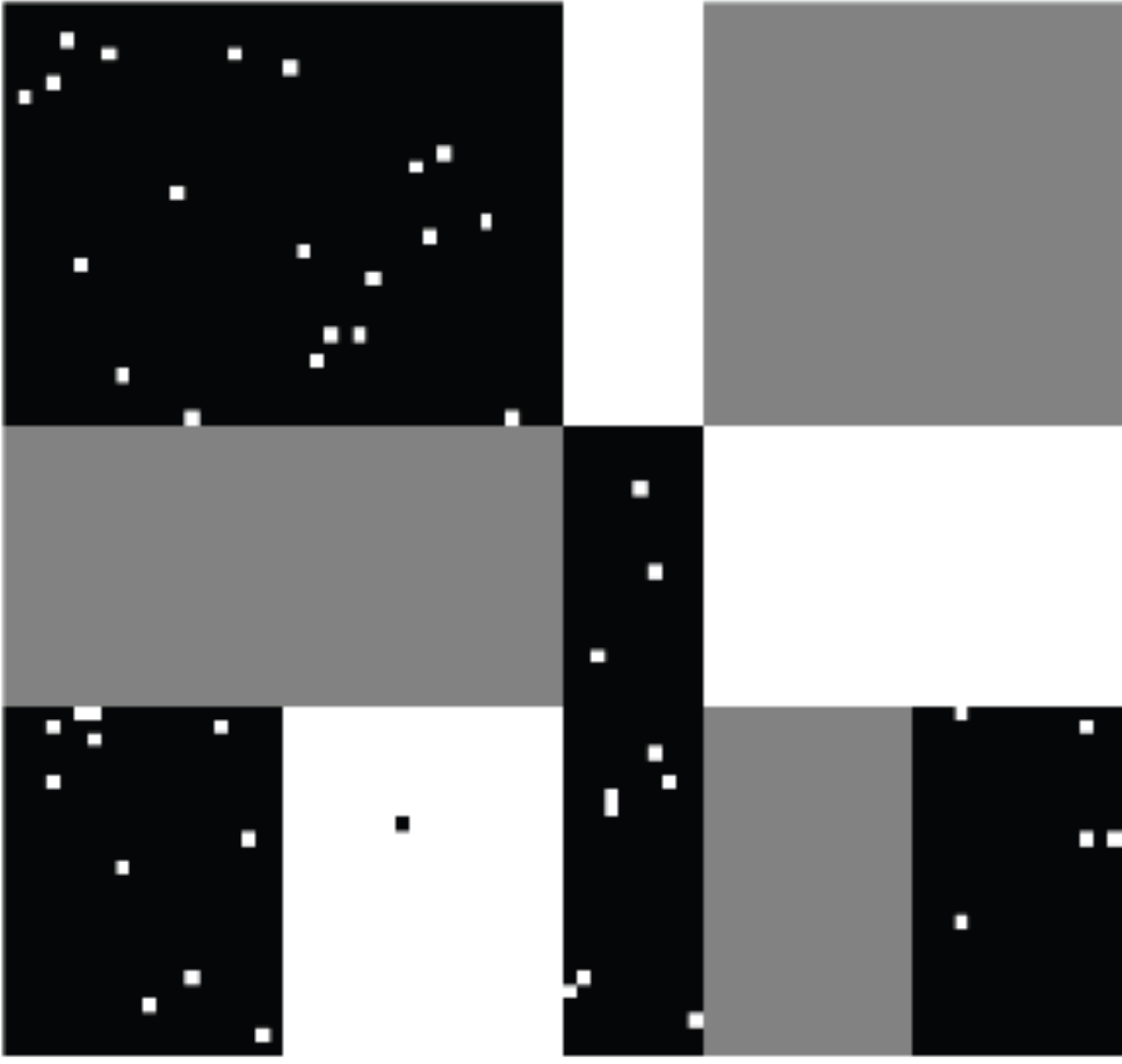}}
\caption{Clustering a synthetic PolSAR image with $k$-means and stochastic distances.} 
\label{ClusterWishart}
\end{figure}

Fig.~\ref{cluster1} presents a simulated PolSAR image  of $75 \times 80$ pixels with ten regions generated from the scaled complex Wishart law and $12$ looks.
The regions have low, intermediate, and high brightness.
The intermediate brightness region has covariance matrix given in~\eqref{matrixEX}, while the largest and smallest brightnesses regions are simulated from the following matrices, respectively:
$$
\left[ \begin{array}{ccc}
962892&19171-3579\textbf{i}&-154638+191388\textbf{i} \\
&56707&-5798+ 16812\textbf{i} \\
&&472251
\end{array} \right]
$$
and 
$$
\left[ \begin{array}{ccc}
32556& 556+787\textbf{i}&24046-27287\textbf{i} \\
&1647&-146-  482\textbf{i} \\
&&61028
\end{array} \right].
$$
These two matrices were observed in~\cite{PolarimetricSegmentationBSplinesMSSP}, in urban and pasture regions, respectively.

Fig.~\ref{cluster2} shows the initial stage in the clustering process, which is quite far from the ideal solution.
Figs.~\ref{cluster3} to~\ref{cluster6} present the results of using the $k$-means algorithm based on stochastic distances.
Notice that all the distances were able to identify clusters accurately with a few spurious spots.

\begin{figure}[htb]
\centering
\subfigure[AIRSAR image~\label{cluster11}]{\includegraphics[width=.38\linewidth]{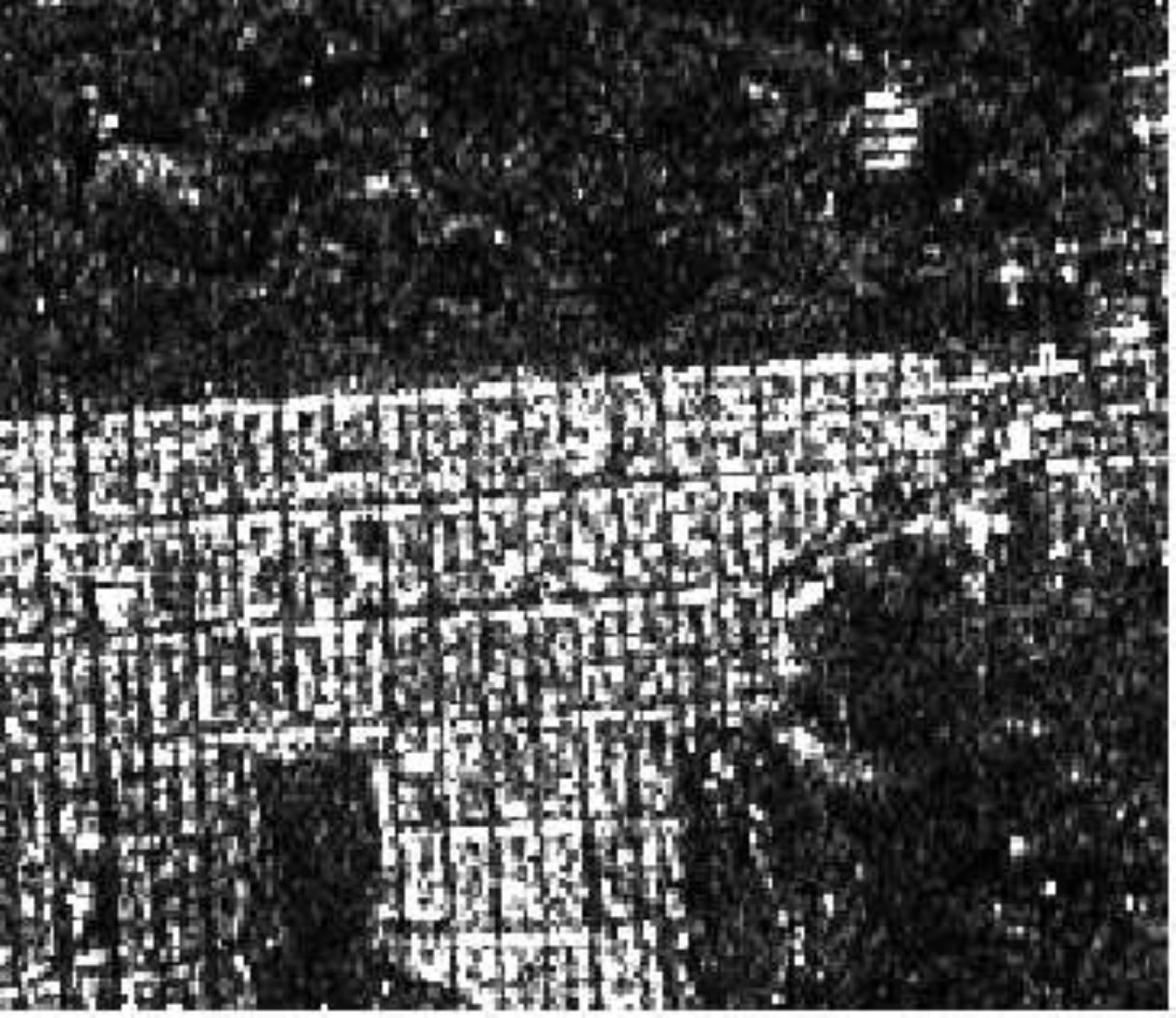}}
\subfigure[Random initialization~\label{cluster22}]{\includegraphics[width=.38\linewidth]{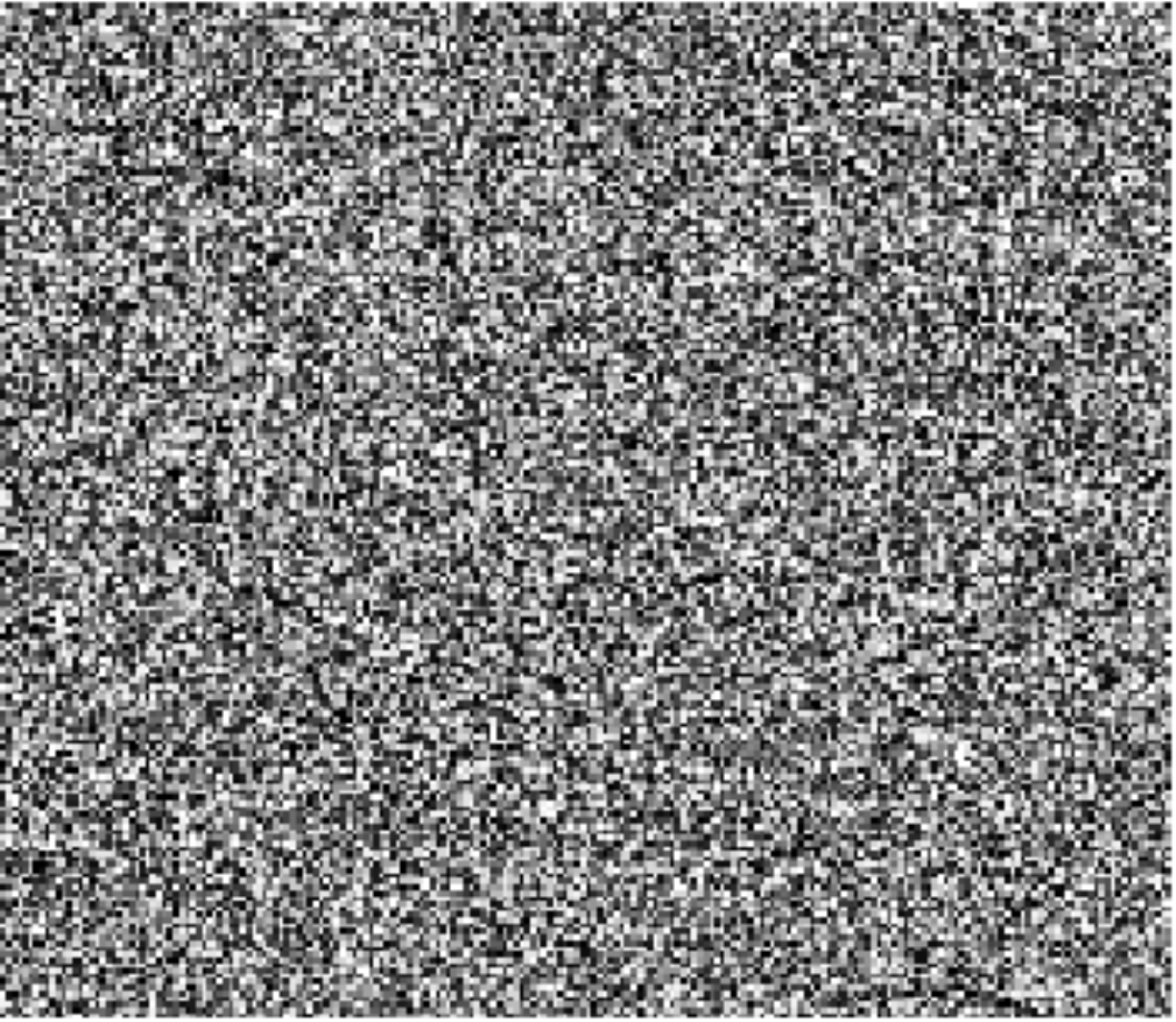}}\\
\subfigure[$d_\text{KL}$ clusters~\label{cluster33}]{\includegraphics[width=.38\linewidth]{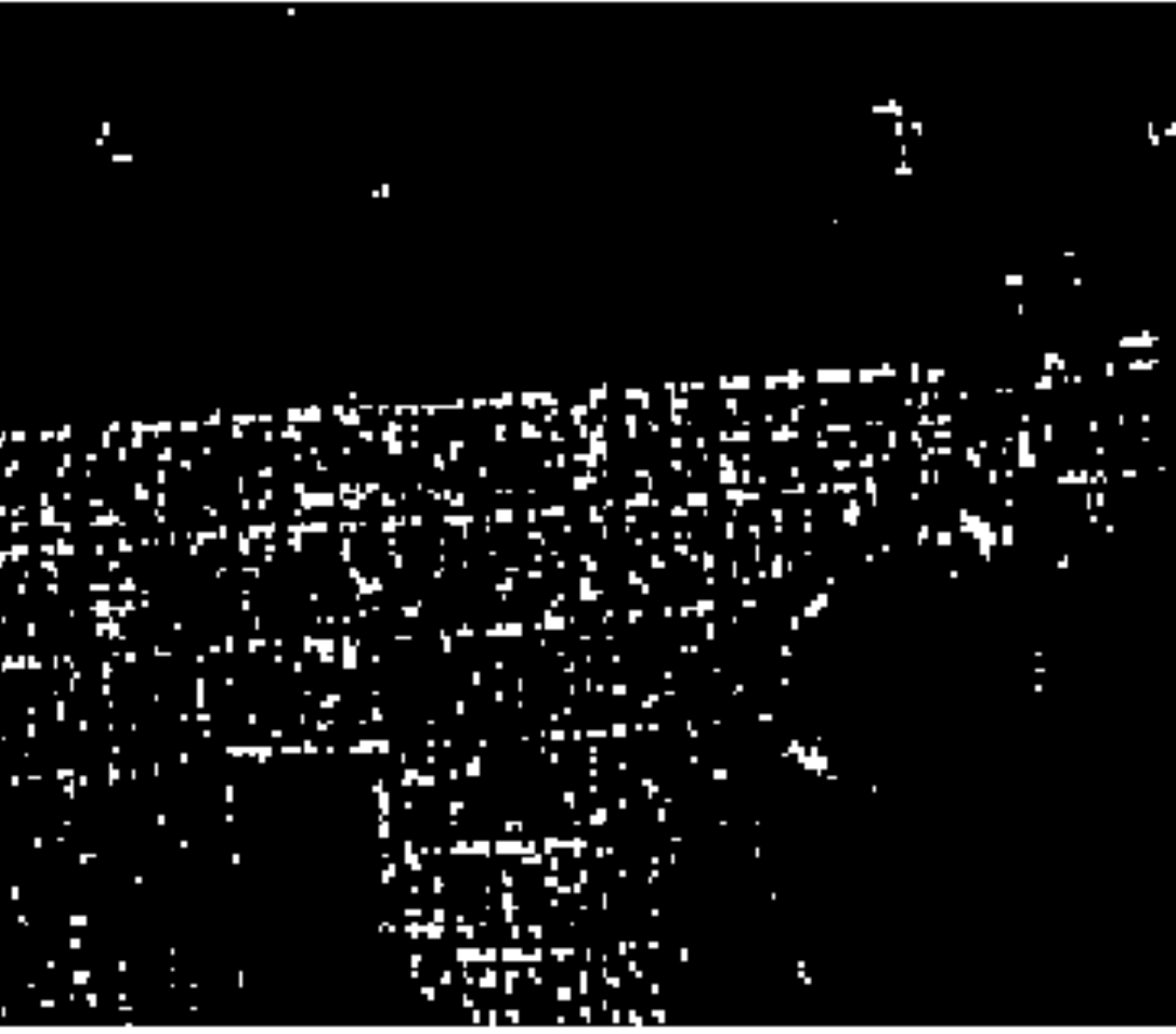}}
\subfigure[$d_\text{B}$  clusters~\label{cluster44}]{\includegraphics[width=.38\linewidth]{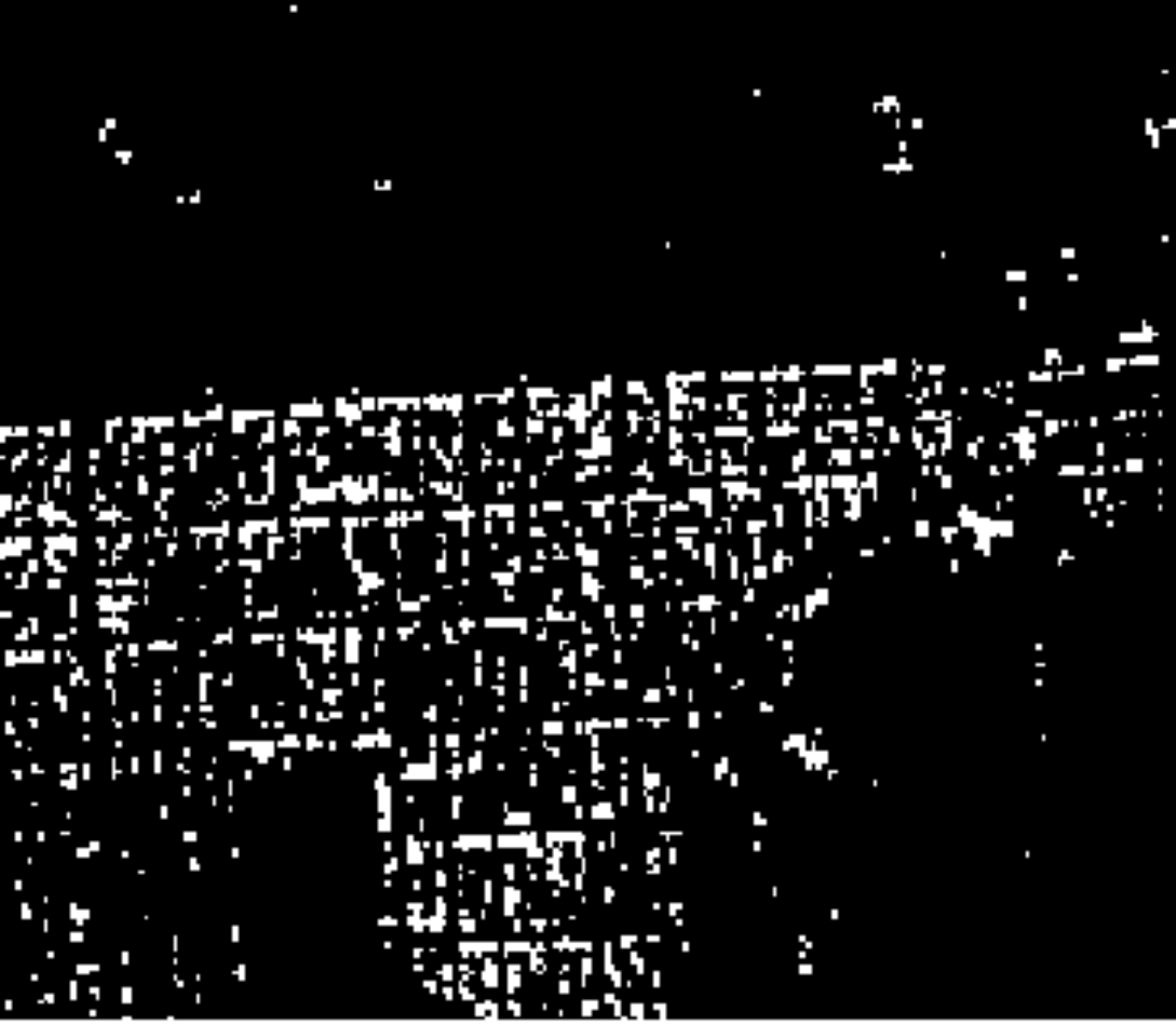}}\\
\subfigure[$d_\text{H}$ clusters~\label{cluster55}]{\includegraphics[width=.38\linewidth]{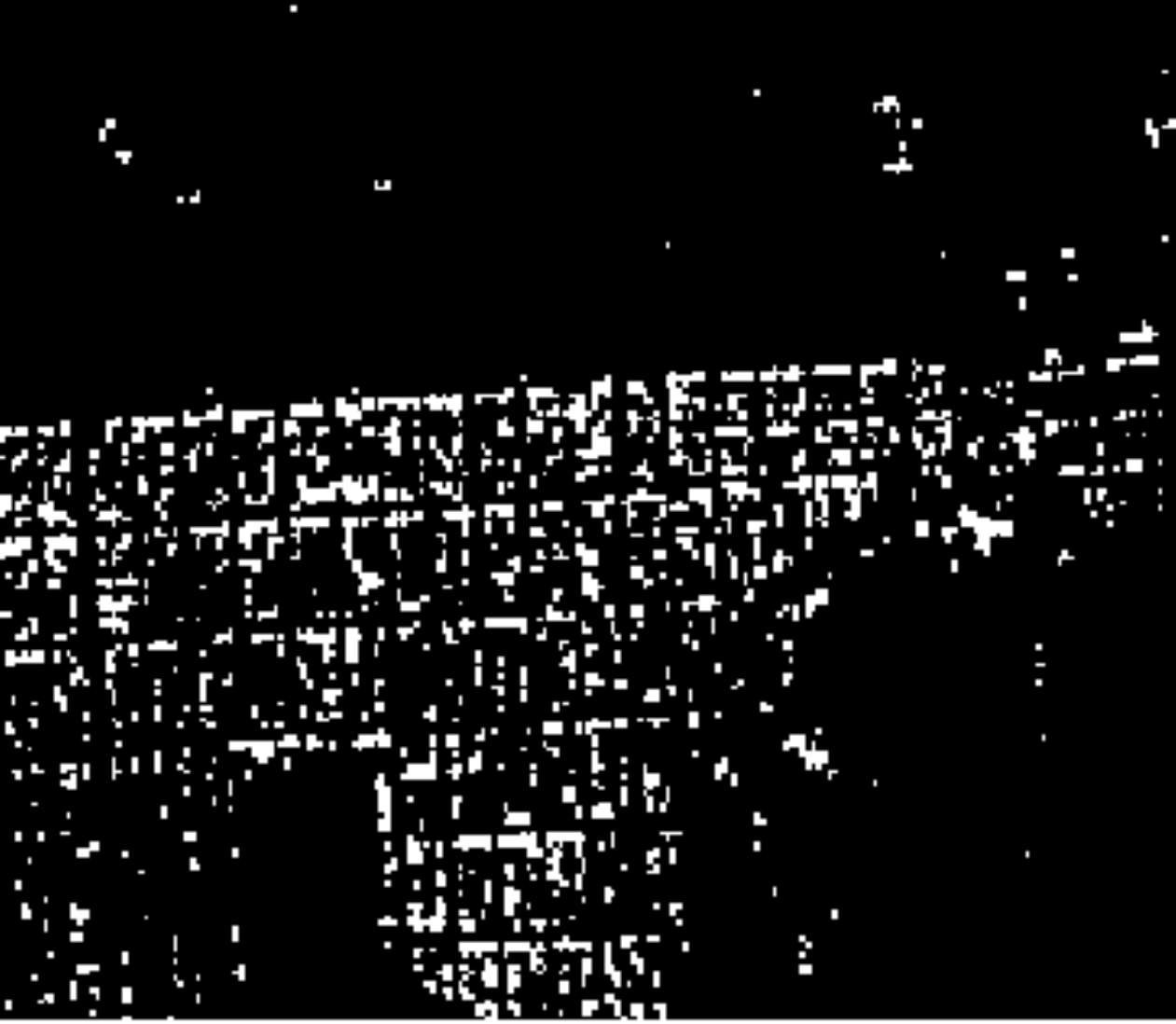}}
\subfigure[$d_\text{R}^{0.1}$ clusters ~\label{cluster66}]{\includegraphics[width=.38\linewidth]{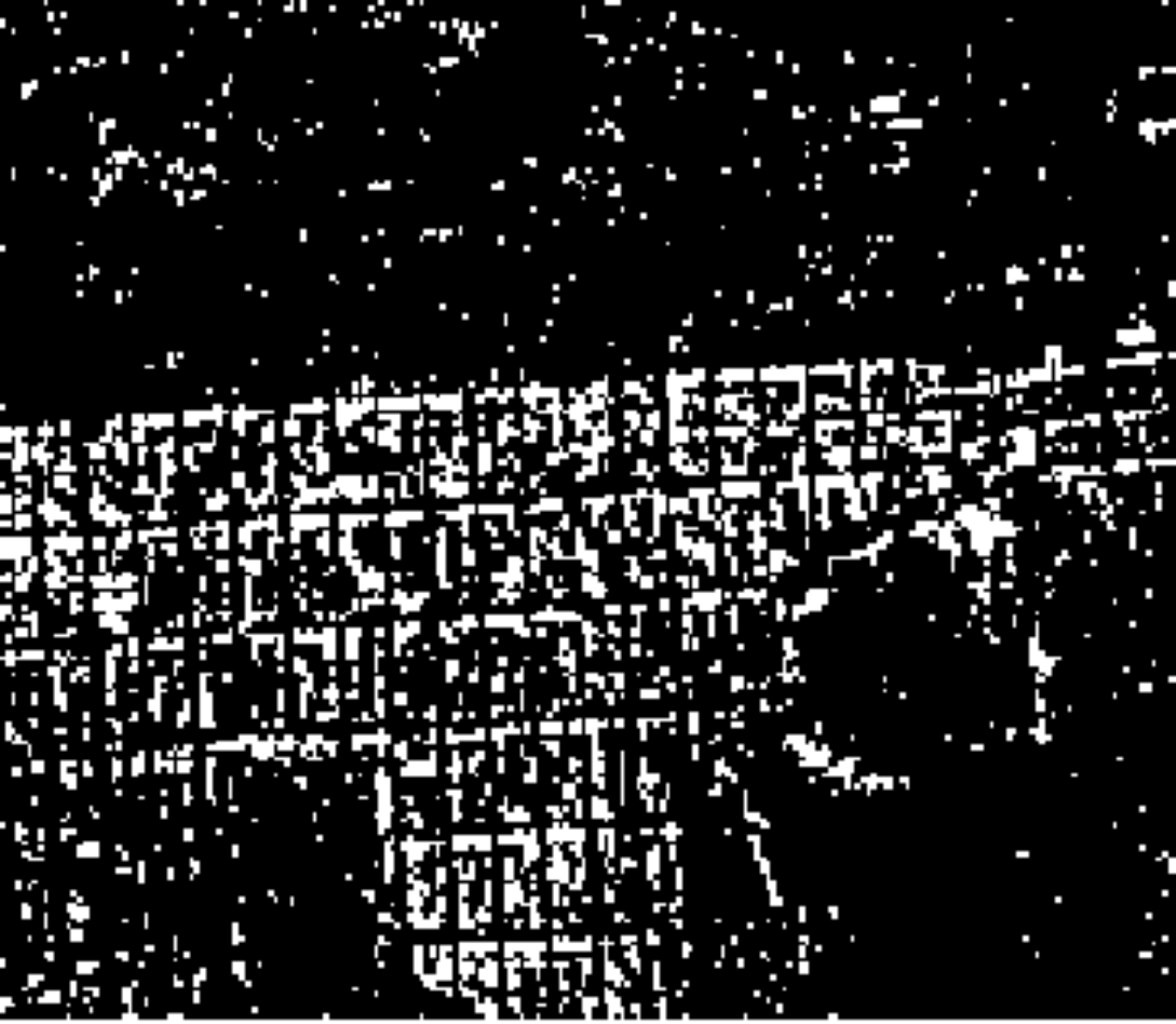}}
\caption{Clustering a PolSAR image with $k$-means and stochastic distances.} 
\label{ClusterWishart1}
\end{figure}

We applied this methodology to a $182\times 210$ pixels area from the San Francisco AIRSAR image (Fig.~\ref{cluster11}).
This area is composed of urban and forest regions. 
Fig.~\ref{cluster22} shows the initial stage of the clustering analysis, which was randomly generated. 
Notice that $d_\text{R}^{0.1}$ gathers more pixels of urban regions than the other distances.
The Kullback-Leibler distance presented the worst performance in terms of the identification of pixels of urban scenarios.
This may be due to the departure from the assumption that a region is Wishart distributed on such situations.

\section{Conclusions}\label{sec:Conclusions}

Analytic expressions of four contrast measures between relaxed complex Wishart distribution were derived for the most general case (different number of looks and different covariance matrices), along with the particular cases of same number of looks and same covariance matrix.
These measures are shown to be scale invariant, and they lead to test statistics with asymptotic $\chi^2$ distribution under the null hypothesis.
Novel inequalities which relate covariance matrices and distances were derived, leading to a new and simple derivation of the revised Wishart and Bartlett distances.
These new expressions can be used in a variety of applications as, for instance, segmentation, and classification.

Those stochastic distances were successfully used as dissimilarities in a $k$-means algorithm.
Data from AIRSAR sensors confirmed the expected behavior of all the distances: distances are smaller when applied to samples of similar roughness, and larger otherwise.

All the proposed statistics based on stochastic distances presented good performance with finite size samples.
In particular, the results provided evidence that the test based on the $d_\text{B}$ has the smallest empirical test size in a variety of situations.
This behavior was confirmed with samples from a PolSAR sensor.

We presented numerical evidence that the statistics based on Hellinger distance overcome the other statistics.
Our results confirm previous studies which pointed the Bartlett distance (a particular case of the Hellinger distance for the same number of looks) as the best option on Wishart distributed data. 
Therefore, the Hellinger test statistics derived from the ($h,\phi$) class of divergences is a reasonable statistical method for assessing if two samples of polarimetric data come from the same distribution.

It is noteworthy that the tests here considered tend to reject more than their nominal levels when dealing with small samples and small number of looks.
Thus, a study of the influence of improved estimators (bias reduction by numerical and analytical approaches, and robust versions, for instance) for the parameters $n$ and $\boldsymbol{\Sigma}$ on the performance of the proposed hypothesis tests is a venue for new research.

Further research will consider models which include heterogeneity~\cite{FreitasFreryCorreia:Environmetrics:03,FreryCorreiaFreitas:ClassifMultifrequency:IEEE:2007,
PolarimetricSegmentationBSplinesMSSP}, robust, improved and nonparametric inference~\cite{AllendeFreryetal:JSCS:05,NonparametricEdgeDetectionSpeckledImagery,SilvaCribariFrery:ImprovedLikelihood:Environmetrics,
VasconcellosFrerySilva:CompStat}, and small samples issues~\cite{FreryCribariSouza:JASP:04}.

\section*{Acknowledgment}

The authors are grateful to CNPq, Facepe, Fapeal, and Capes for their support.

\appendices

\section{The Kullback-Leibler distance in general form}\label{app:KL}

The Kullback-Leibler distance is given by
\begin{align}\label{distKL1}
d_\text{KL}(\boldsymbol{\theta}_1,&\boldsymbol{\theta}_2)= \nonumber\\
&\frac{n_1-n_2}{2}\bigl[\operatorname{E}(\log |\boldsymbol{Z}_1|) -\operatorname{E}(\log |\boldsymbol{Z}_2|)\bigr] \nonumber\\
&-\frac{1}{2}\operatorname{E}\bigl[n_1\operatorname{tr}\bigl(\boldsymbol{\Sigma}_1^{-1}\boldsymbol{Z}_1\bigr)-n_2\operatorname{tr}\bigl( \boldsymbol{\Sigma}_2^{-1}\boldsymbol{Z}_1\bigr)\bigr] \nonumber\\
&+\frac{1}{2}\operatorname{E}\bigl[n_1\operatorname{tr}\bigl(\boldsymbol{\Sigma}_1^{-1}\boldsymbol{Z}_2\bigr)-n_2\operatorname{tr}\bigl( \boldsymbol{\Sigma}_2^{-1}\boldsymbol{Z}_2\bigr)\bigr],
\end{align}
where $\boldsymbol{Z}_i\sim\mathcal{W_R}(\boldsymbol{\Sigma}_i,n_i)$, $i=1,2$. 
According to Anfinsen \textit{et~al.}~\cite{EstimationEquivalentNumberLooksSAR}, we have:
\begin{align}
\operatorname{E}(\log &|\boldsymbol{Z}_i|)=\log |\boldsymbol{\Sigma}_i|+\sum_{k=0}^{p-1}\psi^{(0)}(n_i-k)- p\log n_i\nonumber \\
\mbox{}=&\log |\boldsymbol{\Sigma}_i| + p\psi^{(0)}(n_i-p+1)
+\sum_{k=1}^{p-1}\frac{k}{n_i-k}\nonumber \\
&- p\log n_i ,
\label{KL11}
\end{align}
since $\psi^{(0)}(x+1)=\psi^{(0)}(x)+x^{-1}$, for any $x$ real.
Additionally,
\begin{align}\label{KL12}
\operatorname{E}&[\operatorname{tr}(\boldsymbol{\Sigma}_j^{-1}\boldsymbol{Z}_i)]=
\operatorname{E}\biggl(\sum_{k=1}^p \sum_{\ell=1}^p \delta_{k\ell j}{z}_{k\ell i}\biggr)\nonumber\\ 
&=\sum_{k=1}^p \sum_{\ell=1}^p \delta_{k\ell j}\operatorname{E}({z}_{k\ell i}) 
=\operatorname{tr}[\boldsymbol{\Sigma}_j^{-1}\operatorname{E}(\boldsymbol{Z}_i)] \nonumber\\
&=\left\{ \begin{array}{ll}
         \operatorname{tr}(\boldsymbol{\Sigma}_j^{-1} \boldsymbol{\Sigma}_i )\text{,} & \mbox{if $i\neq j$},\\
        p\text{,} & \mbox{if $i=j$},\end{array} \right.
\end{align} 
where $\delta_{k\ell j}$ and ${z}_{k\ell i}$ are the $(k,\ell)$-th entry of the matrices $\boldsymbol{\Sigma}_j^{-1}$ and $\boldsymbol{Z}_i$, respectively.
Hence, applying~\eqref{KL11} and~\eqref{KL12} into~\eqref{distKL1} yields~(7).

\bibliographystyle{IEEEtran}
\bibliography{BidDistance}

\begin{IEEEbiography}[{\includegraphics[width=1in]{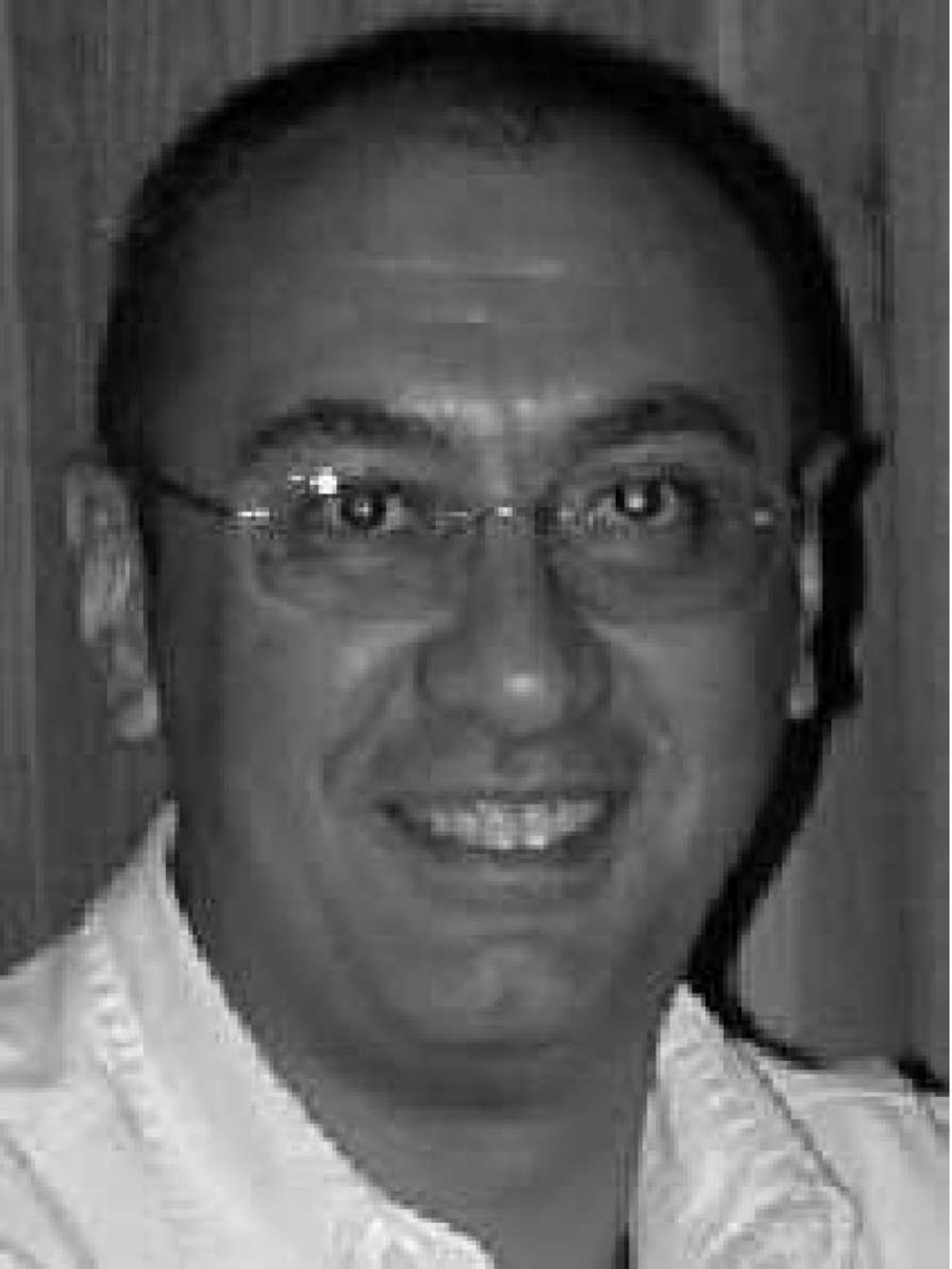}}]{Alejandro C.\ Frery} (S'92--M'95)
received the B.Sc. degree in Electronic and Electrical Engineering from the Universidad de Mendoza, Mendoza, Argentina.
His M.Sc. degree was in Applied Mathematics (Statistics) from the Instituto de Matem\'atica Pura e Aplicada (IMPA, Rio de Janeiro) and his Ph.D. degree was in Applied Computing from the Instituto Nacional de Pesquisas Espaciais (INPE, S\~ao Jos\'e dos Campos, Brazil).
He is currently with the Instituto de Computa\c c\~ao, Universidade Federal de Alagoas, Macei\'o, Brazil.
His research interests are statistical computing and stochastic modelling.
\end{IEEEbiography}

\begin{IEEEbiography}[{\includegraphics[width=1in]{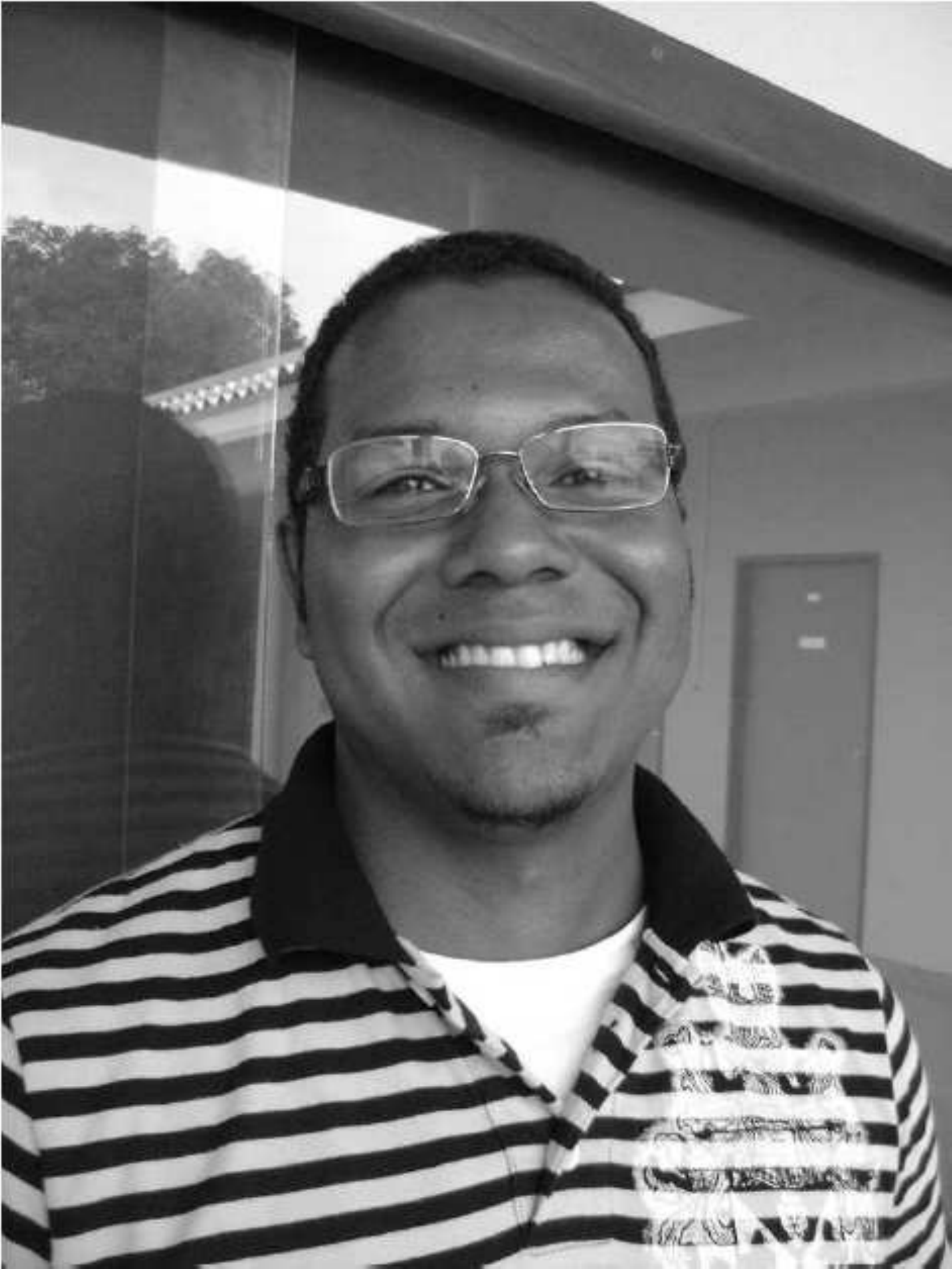}}]{Abra\~ao D.\ C.\ Nascimento}
holds B.Sc.\, M.Sc.\, and D.Sc. degrees in Statistics from Universidade Federal de Pernambuco (UFPE), Brazil, in 2005, 2007, and 2012, respectively.
In 2012, he joined the Department of Statistics at UFPE as Substitute Professor.
His research interests are statistical information theory, inference on random matrices, and asymptotic theory.
\end{IEEEbiography}

\begin{IEEEbiography}[{\includegraphics[width=1in]{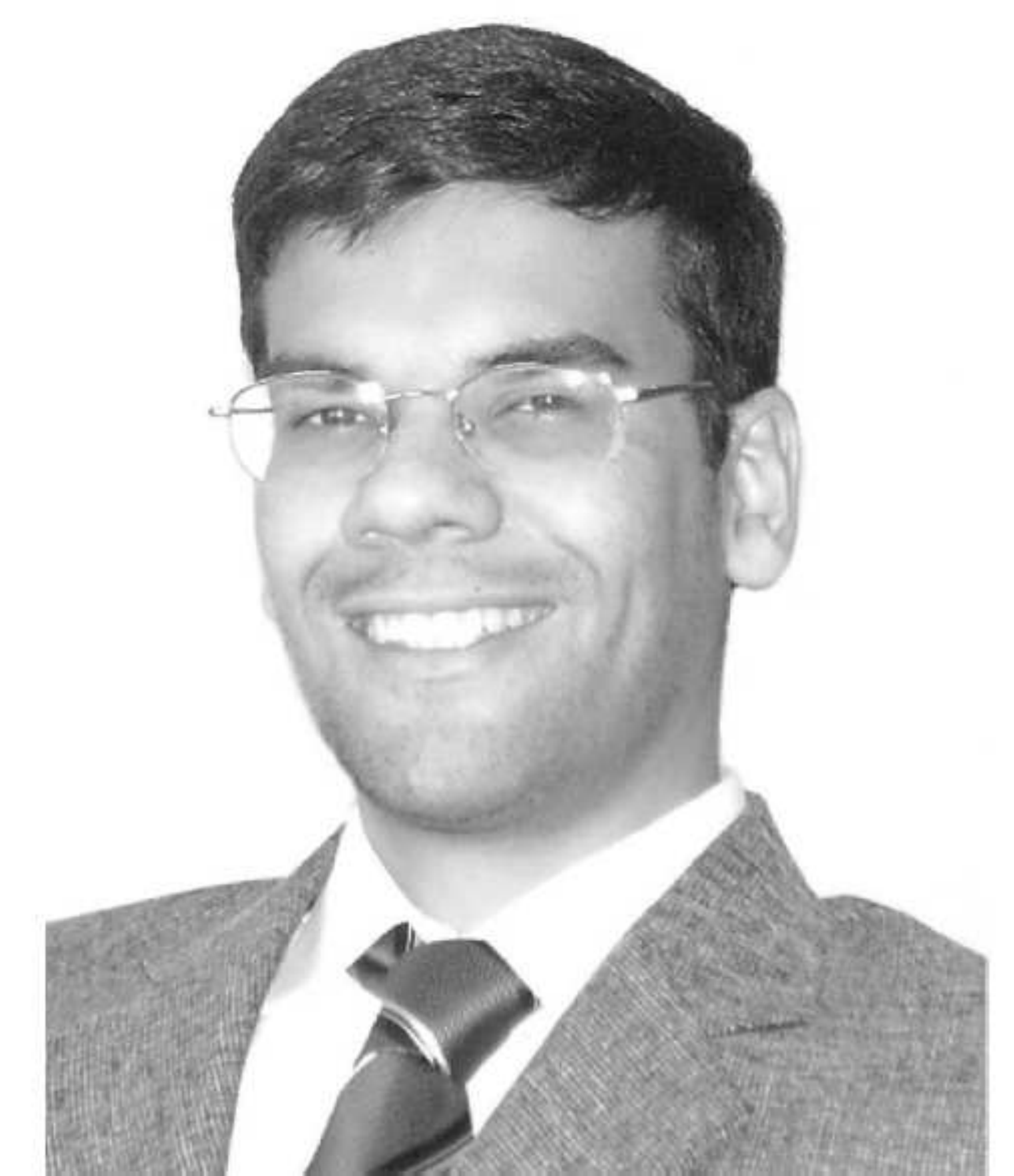}}]{Renato J.\ Cintra} (SM'10)
earned his B.Sc., M.Sc., and D.Sc. degrees
in Electrical Engineering from
Universidade Federal de Pernambuco,
Brazil, in 1999, 2001, and 2005, respectively.
In 2005,
he joined the Department of Statistics at UFPE.
During 2008-2009,
he worked at the University of Calgary, Canada,
as a visiting research fellow.
He is also a graduate faculty member of the
Department of Electrical and Computer Engineering,
University of Akron, OH.
His long term topics of research include
theory and methods for digital signal processing,
communications systems, and applied mathematics.
\end{IEEEbiography}

\vfill

\end{document}